\documentclass[11pt]{article}

\usepackage[preprint]{acl}

\usepackage{times}
\usepackage{latexsym}
\usepackage[T1]{fontenc}
\usepackage[utf8]{inputenc}
\usepackage{microtype}
\usepackage{inconsolata}
\usepackage{graphicx}
\usepackage{subcaption}
\usepackage{booktabs}
\usepackage{amsmath}
\usepackage{amssymb}
\usepackage{mathtools}
\usepackage{amsthm}
\usepackage{tabularx}
\usepackage{longtable}
\usepackage{multirow}
\usepackage{enumitem}
\usepackage{xcolor}
\usepackage{listings}
\usepackage{url}
\usepackage[capitalize,noabbrev]{cleveref}

\theoremstyle{plain}

\theoremstyle{definition}

\theoremstyle{remark}

\title{VideoAgent: All-in-One Framework for Video Understanding and Editing}
\author{
  Hengji Zhou$^{1,2*}$, Lingxuan Huang$^{3*}$, Jian Wang$^4$, Bing Zhou$^4$, 
  Si Wu$^2$, Lianghao Xia$^{1\dagger}$, Chao Huang$^3$ \\
  $^1$Harbin Institute of Technology, Shenzhen \\
  $^2$South China University of Technology \\
  $^3$The University of Hong Kong, $^4$Snap Inc \\
  \texttt{hengjizhou01@gmail.com}, \texttt{xvrhuang@connect.hku.hk}, \\
\texttt{jianwang.cmu@gmail.com}, \texttt{zhoubinwy@gmail.com}, \\
  \texttt{cswusi@scut.edu.cn}, \texttt{aka\_xia@foxmail.com}, \texttt{chaohuang75@gmail.com}
}

\newcommand{\ie}{\textit{i}.\textit{e}.}
\newcommand{\eg}{\textit{e}.\textit{g}.}
\def\model{VideoAgent}

\begin{document}
\maketitle
\footnotetext[1]{$^*$Hengji Zhou and Lingxuan Huang have equal contribution to this work.}
\footnotetext[2]{$^\dagger$Lianghao Xia is the corresponding author.}

\begin{abstract}
Video editing has become essential in digital media creation, yet existing automated systems are restricted to short segment processing and domain-specific tasks. They face two critical limitations: i) inability to handle diverse video comprehension and editing operations, and ii) lack of long-video understanding for coherent narrative creation.
We propose \textbf{\model}, an all-in-one agentic framework addressing these challenges through two key innovations. First, we develop automated video shot creation with shot planning agents for coherent narratives and cross-modal retrieval for aligned visual content. Second, we design a multi-agent orchestration framework integrating over thirty specialized editing agents. Intent parsing filters relevant tools while textual-gradient graph optimization assembles complex editing pipelines.
Extensive experiments on our newly-proposed \textbf{VideoEdit} benchmark and public datasets demonstrate \model's superiority over existing multimodal LLMs and agentic systems. \model\ achieves 87-95\% orchestration success rates while reducing API costs by 60\%. Human evaluation across six video categories shows \model\ produces professional-quality content approaching human-level performance, with ratings only 4\% below human-created videos. We release our code at \url{https://github.com/HKUDS/VideoAgent}.
\end{abstract}

\section{Introduction}
\label{sec:intro}

Video editing has become indispensable in modern digital media creation~\citep{esser2023structure}, as evidenced by the explosive growth of platforms like YouTube and TikTok where millions of creators produce content daily~\citep{chai2023stablevideo}. However, professional video editing requires complex tools, specialized skills, and intricate multi-step workflows that create significant barriers for ordinary users. This gap between widespread content creation demand and technical accessibility has driven the need for automated video editing systems~\citep{wang2025videodirector} that interpret natural language instructions and execute sophisticated editing tasks without professional expertise.

Recent advances in Large Language Models (LLMs)~\citep{minaee2024large} and Vision Language Models (VLMs)~\citep{zhang2024vision} have made it increasingly feasible to leverage AI for multimodal data understanding and editing. Deep Video Discovery (DVD) employs LLM-coordinated tool ensembles to enable autonomous video exploration through systematic "observe-reason-act" cycles~\citep{zhang2025deep}. VideoRAG introduces a dual-channel architecture combining knowledge graphs with multimodal encoding for complex reasoning over long videos~\citep{ren2025videorag}. mPLUG-2 demonstrates modular multimodal foundations using dual visual encoders for unified text, image, and video understanding~\citep{xu2023mplug}. Despite these pioneering efforts, none have adequately addressed the complex requirements of automated video editing. Existing approaches suffer from two critical limitations: first, they focus on specific understanding or editing tasks and cannot handle the diverse range of operations required in real-world scenarios. Second, they lack the long-video understanding and planning capabilities essential for human-level video editing, preventing the creation of videos with sufficient length and coherent narrative structure.

To address these limitations and achieve human-level automated video creation capabilities, we identify two fundamental challenges:
\begin{itemize}[leftmargin=*]
    \item \textbf{Coherent Long-Form Video Planning}. Professional video creation demands coherent narratives spanning multiple shots and extended durations, requiring sophisticated top-down shot planning with multi-step decomposition and strong reasoning capabilities. For instance, creating a movie trailer requires understanding character arcs, identifying climactic moments, and sequencing emotional beats to build narrative tension---a process that involves complex hierarchical planning from story structure down to individual shot specifications. Such shot planning must be conducted with comprehensive awareness of available visual materials, demanding global understanding and precise retrieval across extensive content libraries. Consider the task of creating a highlight reel from a 2-hour film: the system must possess deep comprehension of both granular visual details and overarching narrative elements to select and sequence appropriate clips. Current approaches predominantly focus on isolated segment understanding and editing, lacking the holistic story creation capabilities essential for professional-quality content.
    \item \textbf{Multi-Agent Workflow Orchestration}. Video editing requires orchestrating diverse tools---from foundational utilities (audio extraction, rhythm detection) to sophisticated creative agents (dialogue creation, narrative structuring)---across modalities with interdependencies. Effective production demands systems that can parse natural language instructions, identify creative patterns, and dynamically compose non-linear workflow networks coordinating multi-modal processing. For instance, creating a music video synchronized to beat drops requires seamlessly integrating audio analysis, rhythm detection, visual retrieval, and temporal alignment, where each component's output feeds subsequent operations. Existing methods are constrained to specific domains or predefined workflows, lacking comprehensive tool libraries and adaptive orchestration capabilities for general-purpose creation.
\end{itemize}

\begin{figure}[t]
  \centering
  \includegraphics[width=\linewidth]{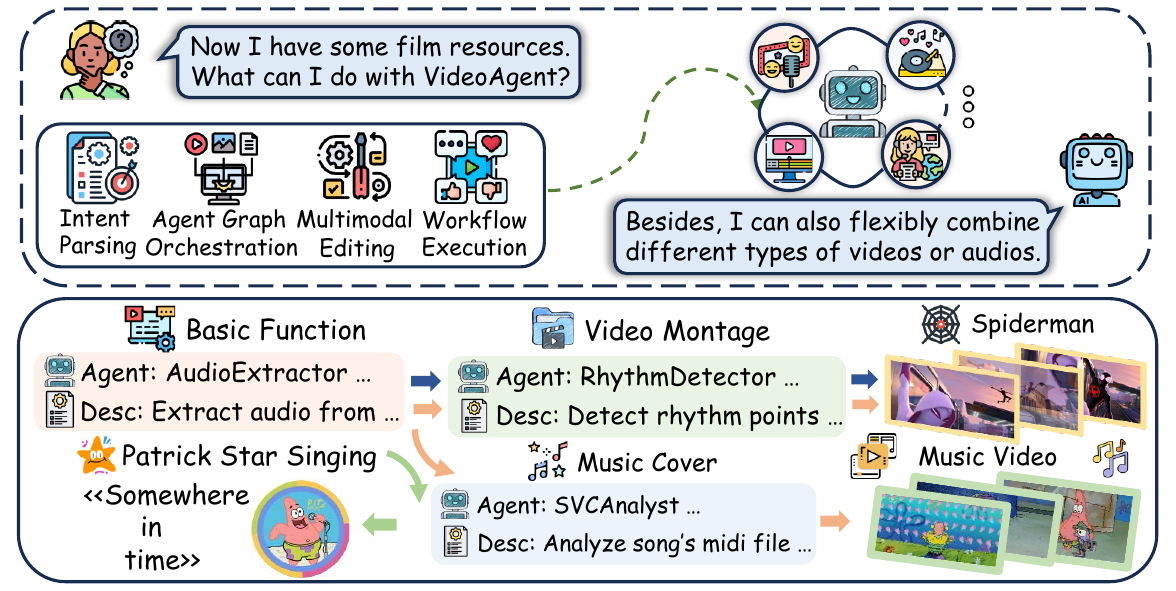}
  \vspace{-0.2in}
  \caption{Automated video editing with \model.} 
  \label{fig:intro}
  \vspace{-0.12in}
\end{figure}

To tackle these challenges, we propose \model, an all-in-one agentic framework for video understanding and editing. Our approach first develops an automated video shot creation system that employs planning agents to generate coherent narrative structures, coupled with cross-modal retrieval mechanisms to extract semantically aligned visual content from extensive material libraries. To enable general-purpose creation across diverse genres, we further design a multi-agent orchestration framework that integrates over thirty specialized editing agents through dynamic workflow composition, where intent parsing mechanisms efficiently filter tools and textual-gradient graph optimization adaptively assembles complex editing pipelines.

Our contributions are summarized as follows:
\begin{itemize}[leftmargin=*]
    \item \textbf{All-in-one Agentic Framework}. We propose \model, a comprehensive agentic framework that unifies video understanding and editing tasks, enabling product-level automated video creation across diverse genres and production workflows.

    \item \textbf{Novel Technical Solutions}. We introduce global-aware shot creation and textual-gradient graph optimization, addressing the coherent long-video planning and multi-agent orchestration challenges that  hinder fully automated video editing.

    \item \textbf{Comprehensive Evaluation}. We construct the VideoEdit benchmark with high-quality instructions and human-aligned evaluation. Experiments show \model's superiority over existing systems, achieving 87--95\% success rates while reducing API costs by 60\%. Human evaluation shows quality ratings only 4\% below professional human-created videos. 

\end{itemize}
\vspace{0.2in}

\vspace{-0.3in}
\section{Methodology}
\label{sec:solution}

\begin{figure*}[t]
  \centering
  \includegraphics[width=\linewidth]{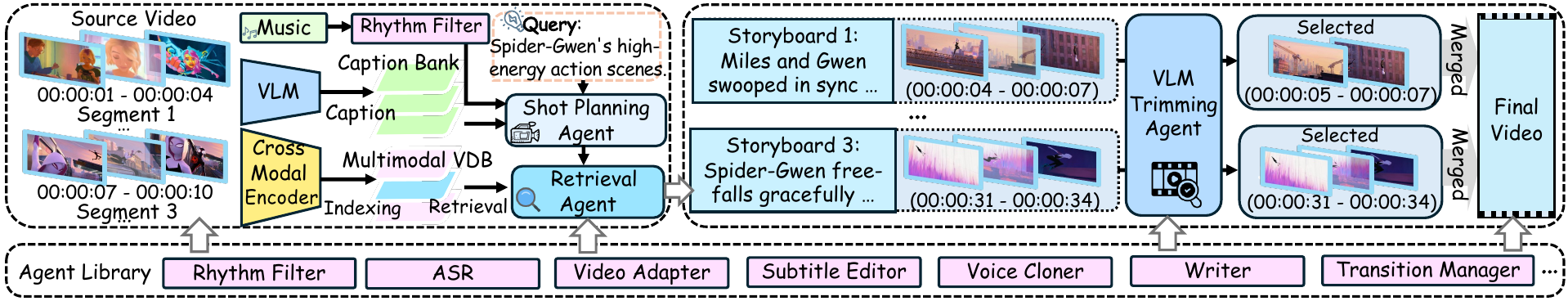}
  \vspace{-0.24in}
  \caption{Automated video shot creation with shot planning, video retrieval and trimming.} 
  \label{fig:framework2.pdf}
  \vspace{-0.16in}
\end{figure*}

\subsection{Video Editing Task Formalization}
We formally define the video editing task as follows. \textbf{Given i)} a natural language instruction $I$ describing the desired video editing objectives, and \textbf{ii)} a collection of multimodal content resources $\mathcal{M} = \{m_1, m_2, \ldots, m_{|\mathcal{M}|}\}$ (\eg, raw video clips, audio tracks, and textual documents for narration), the system \textbf{produces} a coherent and semantically aligned edited video $V$ that satisfies the specified instruction $I$. The output video $V$ comprises a temporally ordered sequence of video shots $V = (v_1, v_2, \ldots, v_{|V|})$, where each shot $v_i$ represents a coherent video segment containing synchronized visual and audio components. In essence, the video editing task is to build:
\begin{align}
    f: (I, \mathcal{M}) \rightarrow V = (v_1, v_2, \ldots, v_{|V|})
\end{align}

\subsection{Automated Video Shot Creation}
\label{sec:shot_creation}

\subsubsection{Shot Planning Agent}
To generate the video shot sequence $(v_1, v_2, \ldots, v_{|V|})$, we develop a shot planning agent that produces structured shot-level storyboards in natural language. Each storyboard provides a detailed textual description specifying the visual content for the corresponding shot, ensuring all shots collectively form a coherent narrative fulfilling the video creation instruction. This process requires global awareness: the shot planning agent must understand both the overall story structure specified by the user instruction $I$ and the available visual resources $\mathcal{M}$. Formally, the shot planning agent operates as follows:
\begin{align}
    S &= (s_1,\ldots,s_{|V|}) = \textsc{SP}(I, C, |V|; p_s), \\
    C &= \text{LLM}(c_i; p_c)_{i=1}^{|\mathcal{M}|}, ~~ c_i = \mathcal{C}(\mathcal{K}(m_i)),
\end{align}
where $S$ is the generated shot sequence, $C$ is the compressed visual context summarizing all available materials, $c_i$ is the caption for the $i$-th visual material obtained by applying a captioning function $\mathcal{C}$ to its keyframe $\mathcal{K}(m_i)$, $p_s$ and $p_c$ are prompts for shot planning and visual context generation respectively, and $|V|$ is the desired number of shots.

The shot planning agent operates through two sequential steps. First, it processes all keyframe captions to generate a compressed visual summary $C$ that captures available video materials. Second, it combines visual summary with the user instruction $I$ to generate the required number of shot descriptions that form the overall narrative structure.

\subsubsection{Shot Visual Content Retrieval}
Given the shot-level storyboards $S$, \model\ employs paired video indexing and retrieval modules to identify and fetch appropriate video clips for each shot in an efficient manner. The indexing module first extracts keyframes from each video segment using the KeyFrames method mentioned above, then employs cross-modal representation methods (\ie, ImageBind~\citep{Girdhar2023ImageBindOE} and CLIP~\citep{radford2021learning}) to generate embeddings for these keyframes within a unified text-visual representation space. For retrieval, \model\ generates embeddings for each shot's storyboard $s_i$ using the same cross-modal representation approach, then selects the visual content with the highest cosine similarity to the storyboard embedding in the current shot. The indexing and retrieval process are defined as follows:
\begin{gather}
r_i = \arg\max\nolimits_{j} \cos(e_j, \mathcal{E}(s_i)), \\
e_i = \mathcal{E}(\mathcal{K}(m_i)), ~~ i = 1, \ldots, |\mathcal{M}|
\end{gather}
where $r_i$ represents the retrieved material index for the $i$-th shot, $e_i = \mathcal{E}(\mathcal{K}(m_i))$ denotes the embedding of the $i$-th material's keyframe extracted by the cross-modal encoder $\mathcal{E}$, and $\cos(\cdot, \cdot)$ denotes cosine similarity. With this method, our visual content retrieval module fetches semantically aligned visual content to form a coherent narrative following the foregoing text-based shot planning.

\subsubsection{Fine-grained Video Trimming}
Although the previous modules fetch semantically aligned visual materials, these raw materials are trimmed to a fixed uniform length. In real creative work, shot lengths vary to match the rhythm of background music and dialogue duration (see Listing~\ref{lst:videoeditor} for shot length decision agents). To address this, \model\ introduces a fine-grained video trimming agent that employs VLMs to select the most suitable sub-segment from the retrieved material $m_{r_i}$ based on the current shot's storyboard text, yielding the final visual material for that shot: \vspace{-0.04in}
\begin{gather}
[t_{\text{start}_i}, t_{\text{end}_i}] = \text{VLM}(m_{r_i}, s_i, t_{i+1}-t_i; p_t), \\
    v_i=m_{r_i}[t_{\text{start}_i}:t_{\text{end}_i}], ~~~~i = 1, 2, \ldots, |V|
\end{gather}
where $v_i$ is the final trimmed video segment for the $i$-th shot, $m_{r_i}$ is the retrieved raw material, $t_{\text{start}_i}$ and $t_{\text{end}_i}$ are the start and end timestamps for trimming, $t_i$ denotes the target start timestamp of the $i$-th shot in the output video timeline (determined by the shot length decision agent, see Listing~\ref{lst:videoeditor}), $t_{i+1}-t_i$ represents the target duration for the shot, and $p_t$ is the prompt for video trimming guidance. Through this adaptive trimming design, the system achieves precise temporal alignment between visual content and narrative requirements. \vspace{-0.12in}
\begin{figure*}[t]
  \centering
  \includegraphics[width=\linewidth]{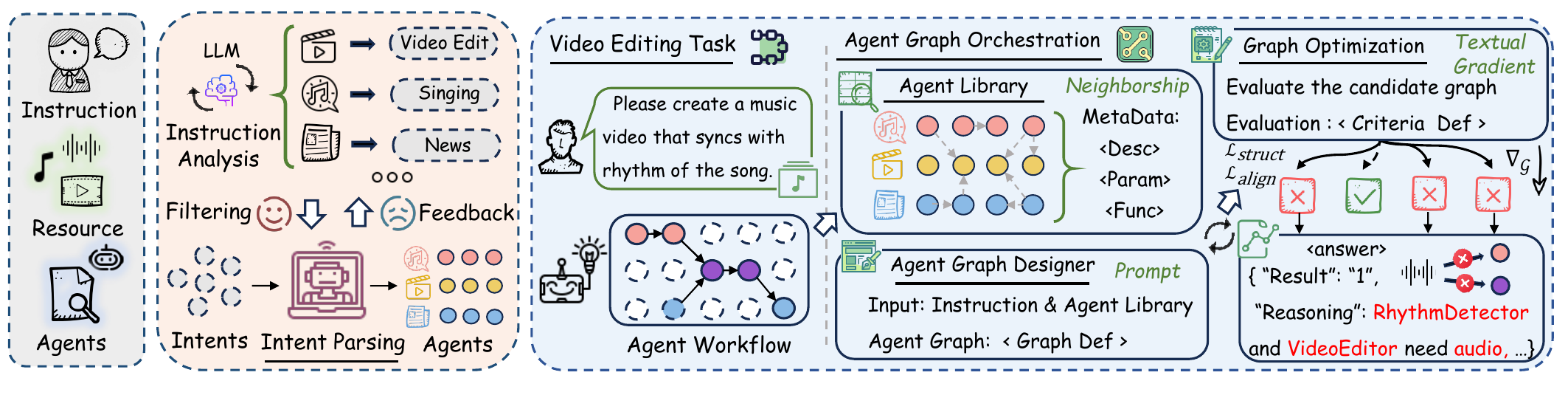}
  \vspace{-0.24in}
  \caption{Video editing with agent graph orchestration and execution.}
  \label{fig:framework}
  \vspace{-0.16in}
\end{figure*}

\subsection{Multi-Agent Graph Orchestration}

\subsubsection{Multimodal Editing Agent Library}
To enhance \model\ with advanced video editing capabilities beyond simple concatenation, we develop a multi-agent system for sophisticated video editing operations. This system enables complex creative transformations requiring deep understanding of content semantics, artistic styles, and technical production workflows. The system encompasses two primary categories of tool agents:
\vspace{-0.25in}
\begin{itemize}[leftmargin=*]
    \item \textbf{Foundational multimodal processing tools} that provide essential technical operations such as music beat extraction, multi-emotional voice synthesis, face swapping, lip-sync generation, audio transcription, and visual style transfer.
    \vspace{-0.05in}
    \item \textbf{Creation prompts with human expertise} that offer guidance for content generation including style-specific scriptwriting, lyrical composition, and meme-based humor creation. For example, the system can transform English stand-up routines into Chinese crosstalk, produce songs with lyrical and vocal adaptations, generate derivative meme content, or synchronize video cuts with dynamic background music rhythms.
\end{itemize}
\vspace{-0.05in}
To support broad spectrum of editing requirements, we develop a comprehensive Agent Library spanning diverse creation genres. Notably, the shot creation functionalities (Section~\ref{sec:shot_creation}) are also wrapped as tool agents, ensuring all capabilities are \textbf{unified within the \model\ framework}. Details of all tool agents are provided in Appendix~\ref{sec:tooluse}.

\subsubsection{Textual-Gradient Graph Optimization}
Conventional agent orchestration relies on single-pass graph generation, where LLMs construct workflows through direct reasoning without systematic quality assessment. We reformulate graph construction as an \textbf{iterative optimization problem over graph topology space}, where structural deficiencies are progressively corrected through textual gradient descent. This enables principled refinement of graph structures guided by explicit quality metrics rather than implicit heuristics.

Formally, let $A=\{A_1, A_2, \ldots, A_n\}$ denote the agent library. An agent graph is a directed acyclic graph $\mathcal{G}=(\mathbb{A}, E)$ where $\mathbb{A} \subseteq A$ is a subset of selected agents and $E \subseteq \mathbb{A} \times \mathbb{A}$ encodes directed inter-agent dependencies. Given instruction $I$, an intent parsing mechanism extracts a set of required intents $\mathcal{T}$ that the agent graph must fulfill.

\noindent\textbf{Optimization Objective.}
We seek an optimal agent graph $\mathcal{G}^*$ that minimizes construction error under instruction $I$ and structural constraints:
\begin{align}
\label{eq:graph-objective}
\mathcal{G}^* = \underset{\mathcal{G} \in \mathbb{G}}{\arg\min}  \left[ \mathcal{L}_{\text{struct}}(\mathcal{G}) + \lambda \mathcal{L}_{\text{align}}(\mathcal{G}, I) \right],
\end{align}
where $\mathbb{G}$ denotes the space of all valid directed acyclic graphs over agent library $A$, $\mathcal{L}_{\text{struct}}$ measures structural validity violations (\eg, cyclic dependencies, disconnected components), $\mathcal{L}_{\text{align}}$ quantifies instruction-graph alignment error, and $\lambda$ balances structural and semantic objectives. The structural loss is formally defined as:
\begin{align}
\mathcal{L}_{\text{struct}}(\mathcal{G}) = \alpha(1 - \tau(\mathcal{G})) + \beta \frac{|\Gamma(\mathcal{G})| - 1}{|A|},
\end{align}
where $\tau$ is the acyclicity indicator returning 1 if $\mathcal{G}$ admits a valid topological ordering and 0 otherwise, $\Gamma(\mathcal{G})$ denotes the set of connected components in $\mathcal{G}$, and $\alpha, \beta$ weight the acyclicity and connectivity penalties. The first term penalizes cycles that break workflow executability, while the second penalizes fragmentation into disconnected subgraphs.

\noindent\textbf{Graph Quality Assessment.}
At iteration $t$, we evaluate the graph $\mathcal{G}^{(t)}$ through multi-faceted quality metrics. The assessment examines: (i) topological properties via acyclicity validation $\tau$ ensuring valid execution orderings, (ii) intent parsing coverage by comparing extracted intents $\mathcal{T}$ against agent capabilities, and (iii) edge consistency through compatibility scoring between connected agents. The quality metrics can be defined as follows:
\begin{align}
\chi = \mathbb{E}&_{ij\in E}\mathbb{I}[\Psi_{ij}\neq\emptyset], ~
\mathcal{Q}^{(t)} = (\tau,\kappa,\chi)^{(t)}, \\
\kappa(\mathcal{T}) &= |\mathcal{T} \cap \textstyle\bigcup_{A_i \in \mathbb{A}} \Psi(A_i)| \cdot |\mathcal{T}|^{-1}
\end{align}
where $\chi$ measures edge compatibility ratio, averaged over all edges $(A_i,A_j)\in E$ in the graph, $\kappa$ measures the intent coverage ratio, $\Psi_{ij} = \Psi(A_i) \cap \Psi(A_j)$ denotes shared capabilities between connected agents, and $\mathbb{I}[\cdot]$ is the indicator function. $\Psi: A \to 2^{\mathcal{I}}$ maps each agent to its executable capabilities over intent space $\mathcal{I}$, $\mathbb{A}$ is the set of agents in $\mathcal{G}^{(t)}$, and $\mathcal{Q}^{(t)}$ is the composite quality signal at iteration $t$. This graph quality assessment produces a \textbf{structured feedback signal} identifying specific graph deficiencies requiring correction.

\noindent\textbf{Textual Gradient for Graph Topology.}
To overcome the incompatibility between LLM-driven heuristic refinement and discrete combinatorial graph topologies, inspired by recent works in agent-based optimization~\citep{hu2024adas, zhang2025multi}, we introduce \textbf{graph-structured textual gradients} that encode topology modifications in natural language. The LLM synthesizes gradient directions by reasoning over: (i) current graph structure $\mathcal{G}^{(t)}$ and quality metrics $\mathcal{Q}^{(t)}$, (ii) instruction requirements $I$ and intent decomposition $\mathcal{T}$, (iii) agent library $A$ specifying capabilities. Formally, the graph update is defined as follows:
\begin{equation}
\label{eq:graph-update}
\mathcal{G}^{(t+1)} = \mathcal{G}^{(t)} \otimes \nabla_{\mathcal{G}}^{\text{text}} \mathcal{L}(\mathcal{G}^{(t)}),
\end{equation}
where $\otimes$ denotes the graph transformation operator, and the textual gradient is computed:
\begin{equation}
\nabla_{\mathcal{G}}^{\text{text}} \mathcal{L} = \text{LLM}\big(\mathcal{G}^{(t)}, \mathcal{Q}^{(t)}, I, \mathcal{T}, A; p_{\text{topo}}\big),
\end{equation}
where $p_{\text{topo}}$ instructs the LLM to generate graph modifications by: (a) removing edges causing cycles or invalid dependencies, (b) inserting missing agents to cover unaddressed intents, (c) reordering execution sequences to satisfy data flow constraints. This approximates the discrete gradient $\nabla_{\mathcal{G}} [\mathcal{L}_{\text{struct}}(\mathcal{G}) + \lambda \cdot \mathcal{L}_{\text{align}}(\mathcal{G}, I)]$ through symbolic reasoning over discrete graph structures, analogous to combinatorial optimization but realized entirely through natural language transformations.

\noindent\textbf{Convergence Criterion.}
Optimization terminates when: (1) both losses vanish, \ie:
\begin{equation}
\mathcal{L}_{\text{struct}}(\mathcal{G}^{(t+1)}) = 0 \;\land\; \mathcal{L}_{\text{align}}(\mathcal{G}^{(t+1)}, I) = 0,
\end{equation}
or (2) iteration count reaches maximum $T_{\max}$. By recasting agent orchestration as graph-structured textual gradient descent, our method transitions from \textbf{generative graph construction}, where graphs emerge from single LLM invocations to \textbf{optimization-driven refinement} grounded in explicit structural and semantic objectives.

\section{Evaluation}
\label{sec:eval}
We evaluate \model\ through five research questions: \textbf{RQ1} overall performance against baselines; \textbf{RQ2} ablation of key components; \textbf{RQ3} hyperparameter sensitivity; \textbf{RQ4} LLM-human judgment consistency; \textbf{RQ5} real-world case study.

\subsection{Experimental Settings}
\noindent\textbf{Datasets and Evaluation Protocols}. We evaluate \model\ on video understanding and workflow orchestration using two datasets: \textbf{Shot2Story}~\citep{han2023shot2story20k} for video retrieval evaluation, and our new \textbf{VideoEdit} benchmark for workflow orchestration with high-quality video creation instructions and human-aligned judge evaluation across diverse video scenarios. See Appendix \ref{sec:data_protocol} for details.

\noindent \textbf{Baseline Methods}. \model\ is compared with a comprehensive list of baseline methods, including \textbf{i) Pure-Language Agentic Systems:} Chain-of-Thought~\citep{wei2022chain}, LLM Debate~\citep{du2023improving}, Step Back~\citep{zheng2023take}, ExpertPrompting~\citep{xu2023expertprompting}, Intelligent Go-Explore\cite{lu2024intelligent}, GPTSwarm~\citep{zhuge2024gptswarm}, 
Flow~\citep{niu2025flow}, 
GraphCounselor~\citep{gao2025graph} \textbf{ii) Multimodal Understanding LLMs and Agents:} Qwen2.5-VL~\citep{bai2025qwen2}, VideoRAG~\citep{ren2025videorag}, TeaserGen~\citep{TeaserGen}, VideoMind~\citep{liu2025videomind} \textbf{iii) Video Generation Agents:} NoteBookLM~\citep{notebooklm}, Director~\citep{director}, FunClip~\citep{funclip}, NarratoAI~\citep{narratoaI}. Details about baselines are provided in Appendix~\ref{sec:baseline}.

\textbf{Implementation Details} of our \model\ and baseline methods are provided in Appendix~\ref{sec:implement}.

\begin{table*}[t]
\centering
  \small
  \setlength{\tabcolsep}{3.4mm}
  \caption{Workflow orchestration performance comparison in terms of \textit{\textbf{Success Rate}}.  Standard deviations are computed over five independent runs.}
  \label{tab:overall_performance}
\begin{tabular}{
    c|cc|cc|cc|cc
  }
    \hline
    \text{Backbone} & \multicolumn{2}{c|}{\text{Claude-Sonnet-4}} & \multicolumn{2}{c|}{\text{Claude-Sonnet-3.7}} & \multicolumn{2}{c|}{\text{GPT-4o}} & \multicolumn{2}{c}{\text{Deepseek-v3}} \\
    \hline
    \text{Data} & \text{Audio} & \text{Video} & \text{Audio} & \text{Video} & \text{Audio} & \text{Video} & \text{Audio} & \text{Video}\\
    \hline\hline
    \text{CoT} & 0.78 & 0.77 & 0.80 & 0.72 & 0.68 & 0.65 & 0.69 & 0.65\\
    \text{Debate} & 0.66 & 0.52 & 0.67 & 0.66 & 0.76 & 0.69 & 0.61 & 0.50\\
    \text{Step Back} & 0.79 & 0.85 & 0.68 & 0.77 & 0.63 & 0.38 & 0.72 & 0.70 \\
    \text{ExpertPrompting} & 0.71 & 0.76 & 0.78 & 0.73 & 0.69 & 0.68 & 0.75 & 0.70  \\
    \text{Intelligent Go-Explore} & 0.69 & 0.63 & 0.70 & 0.75 & 0.88 & 0.86 & 0.58 & 0.53  \\
    \text{Flow} & 0.62 & 0.64 & 0.84 & 0.83 & 0.68 & 0.60 & 0.66 & 0.61 \\
    \text{GPTSwarm} & 0.69 & 0.83	& 0.68 & 0.86 & 0.64 & 0.80 & 0.73	& 0.81\\
    \text{GraphCounselor} & 0.83 & 0.82 & 0.85 & 0.81 & 0.86 & 0.84 & 0.82 & 0.84\\
    \hline
    \multirow{2}{*}{\text{\model}} & \textbf{0.93} & \textbf{0.87} & \textbf{0.95} & \textbf{0.93} & \textbf{0.90} & \textbf{0.88} & \textbf{0.92} & \textbf{0.89} \\
    & $\pm$0.02 & $\pm$0.01 & $\pm$0.03 & $\pm$0.02 & $\pm$0.02 & $\pm$0.01 & $\pm$0.02 & $\pm$0.02 \\
    \hline
  \end{tabular}
\end{table*}

\begin{table*}[t]
\centering
  \small
  \setlength{\tabcolsep}{0.9mm}
  \caption{Video understanding and retrieval performance comparison in terms of Recall@1 (\%), Embedding Matching score (EM) (\%), Intersection over Union (IoU) (\%), API calling Costs and Time Efficiency (s).}
  \label{tab:video_scene_retrieval}
  \begin{tabular}{
    c|c|c|c|c|c||c|c|c|c|c|c
  }
    \hline
    \text{Method} & \text{Recall} & \text{EM} & \text{IoU} & \text{Cost} & \text{Time} & \text{Method} & \text{Recall} & \text{EM} & \text{IoU} & \text{Cost} & \text{Time} \\
    \hline
    \hline
    \text{Claude-Sonnet-3.7} &  \textbf{46.03} &  27.95 &   23.91 & 0.374 & 43 
    & \text{Ours-Claude-Sonnet-3.7} & 44.27& \textbf{28.18} & \textbf{24.81} & 0.147 & 37 \\
    
    \text{Claude-Sonnet-3.5} & 27.28 & 27.35 & 12.72 & 0.375 & 41 
    & \text{Ours-Claude-Sonnet-3.5} & \textbf{38.70} & \textbf{28.45} & \textbf{21.32} & 0.147 & 36 \\

    \text{Gemini-2.5-pro} &  45.98 &  27.78 &  \textbf{25.91}  & 0.349 & 42
    & \text{Ours-Gemini-2.5-pro} & \textbf{47.24} & \textbf{28.21} & 25.74 & 0.136 & 37\\
    
    \text{Gemini-2.5-flash} & 30.04 & 28.04 & 16.69 & 0.070 & 39
    & \text{Ours-Gemini-2.5-flash} &  \textbf{44.93} & \textbf{28.25} & \textbf{25.07} & 0.028 & 36\\
    
    \text{GPT-4o} & 31.22 & 27.64 & 18.53 & 0.253 & 43
    & \text{Ours-GPT-4o} & \textbf{48.85} & \textbf{28.26} & \textbf{26.99} & 0.099 & 36\\
    \hline
    \text{VideoRAG} & 31.03 & 15.84 & 14.35  & 0.100 & 67
    &\text{Qwen-2.5-VL-72B-Instruct} & 18.89 & 27.99 & 10.51 & - & 35 \\
    \text{VideoMind-7B} & 38.26 & 27.75 & 19.67 & - & 21 &
    \text{TeaserGen} & 41.88 & 27.62 & 20.91 & 0.079 & 25 \\
    \hline
  \end{tabular}
\end{table*}

\subsection{Overall Performance Comparison (RQ1)}

We first compare \model\ with baselines on orchestrating effective video editing workflows and video retrieval. As results shown in tables~\ref{tab:overall_performance} and~\ref{tab:video_scene_retrieval}, we observe the following: \textbf{Video Editing Orchestration.} \model\ consistently outperforms baselines across all backbones, achieving success rates of 0.87--0.95 and surpassing the best baseline by 2--25\%. The flexible agent graph and gradient-based orchestration effectively handle the multimodal complexity that existing agentic methods struggle with. \textbf{Video Understanding.} \model\ substantially improves retrieval across foundational models---GPT-4o's Recall rises from 31.22\% to 48.85\%, Gemini-2.5-flash from 30.04\% to 44.93\%---via the shot planning agent's global perception that generates coherent storyboards for accurate retrieval, outperforming RAG and agent frameworks in visual-language alignment. \textbf{Cost-Efficiency.} \model\ reduces API costs by 60\% while maintaining superior performance, as strategic shot planning and targeted retrieval minimize redundant processing compared to baselines that feed all candidate shots into context.

To further verify \model's generalization capability and computational efficiency, we conduct additional experiments on VideoRepurpose~\citep{Wu2024VideoRF} and a workflow orchestration efficiency study in Appendices \ref{sec:videorepurpose_eval} and \ref{sec:workflow_efficiency}.

\subsubsection{Human Evaluation on Video Quality}
We evaluate video quality via human ratings from 26 participants on 18 trending videos (>1M views each), scored on consistency, audio quality, and scene diversity (details in Appendix~\ref{sec:human evaluation}). Figure~\ref{fig:video_quality} compares \model\ against four baselines and human-created content. \textbf{Outperforming Baselines.} \model\ succeeds across all categories where existing methods partially fail, achieving significantly higher scores through effective tool orchestration and accurate visual understanding. \textbf{Matching Human Creators.} \model\ approaches or exceeds human-created videos in several categories, delivering consistently high quality while human creators exhibit more variable skill levels.

\subsection{Ablation Study (RQ2)}

\begin{figure}[t]
  \centering
  \includegraphics[width=0.45\textwidth]{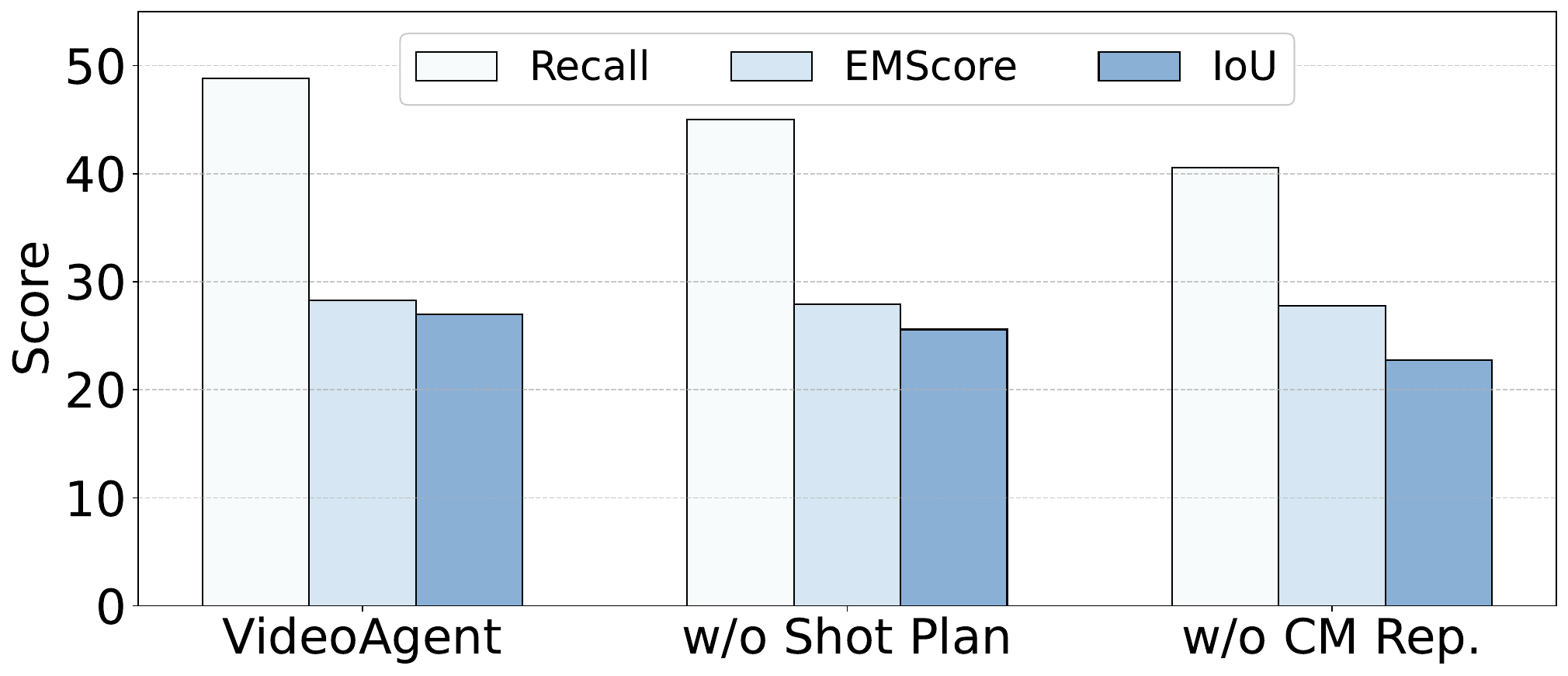}
  \vspace{-0.05in}
  \captionof{figure}{\model\ ablation study results.}
  \label{fig:ablation_comparison}
  \vspace{-0.2in}
\end{figure}

We ablate key components on both video shot creation (Figure~\ref{fig:ablation_comparison}) and workflow orchestration (Table~\ref{tab:ablation1}). \textbf{Shot Planning Agent.} Removing the planner and using raw queries directly causes significant retrieval drops across all metrics, confirming the necessity of global shot planning even with user-provided descriptions. \textbf{Cross-Modal Representation.} Replacing our CM Rep. with caption-then-encoding via a text embedding model yields inferior results, validating that our cross-modal paradigm better preserves visual fidelity. \textbf{Agent Graph Orchestration.} Removing gradient-based optimization (-G) cuts success rates from >90\% to below 55\%, highlighting its critical role in handling complex video editing workflows. \textbf{Intent Parsing.} While its removal (-I) has limited accuracy impact, costs increase notably, confirming its cost-efficiency. \textbf{Key Agent Dependencies.} Ablating individual agents (LoudnessNormalizer (LN), AudioExtractor (AE), StandUpSynth (SS)) reveals proportional performance drops correlated with graph centrality, demonstrating effective multi-agent orchestration even in complex scenarios.

\begin{figure}[t]
  \centering
  \includegraphics[width=0.45\textwidth]{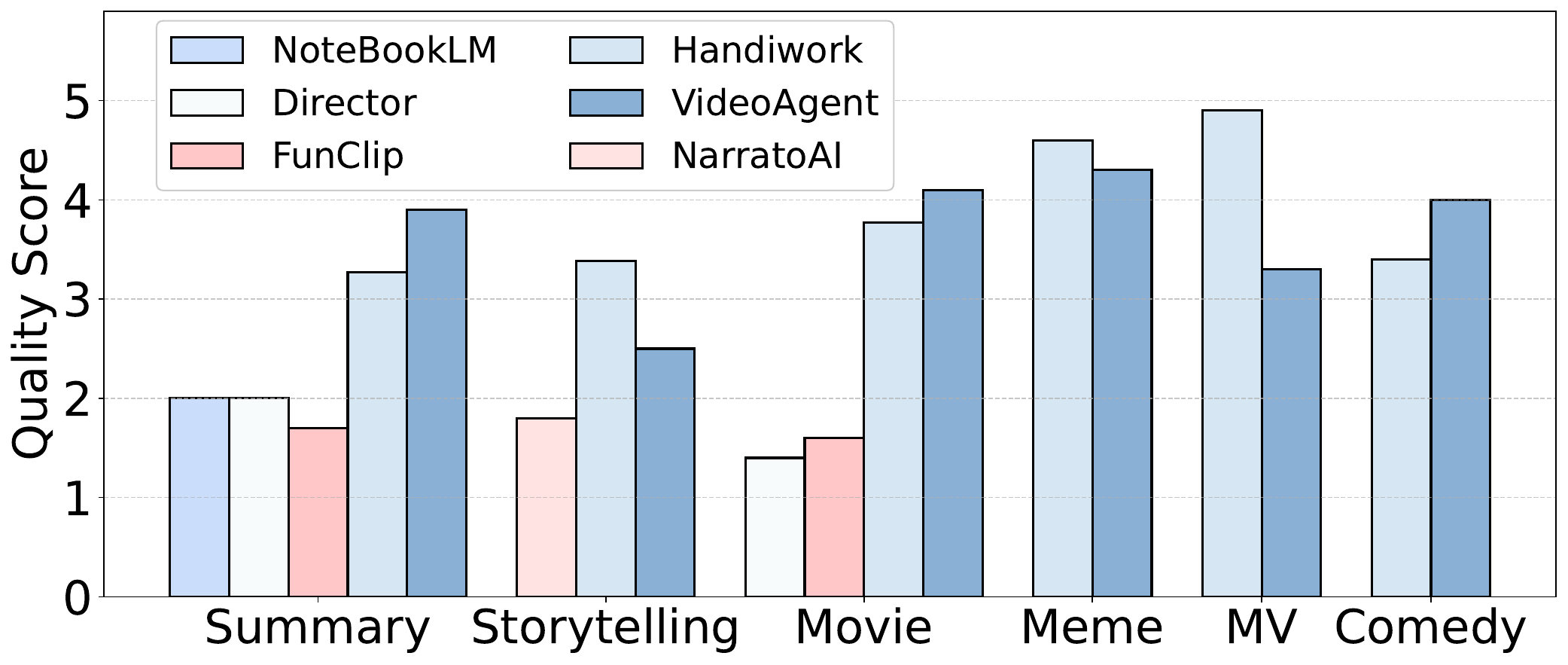}
  \vspace{-0.05in}
  \captionof{figure}{Human-rated video quality assessment.}
  \label{fig:video_quality}
  \vspace{-0.1in}
\end{figure}

\begin{table}[t]
  \centering
  \small
  \setlength{\tabcolsep}{3.5pt}
  \caption{Performance of different ablated \model, in terms of Success Rate (SR) and Cost (\$).}
  \label{tab:ablation1}
  \vspace{-0.1in}
  \begin{tabular}{
      c| cccc|cccc
  }
  \hline
  Backbone & \multicolumn{4}{c|}{Claude-Sonnet-3.7} & \multicolumn{4}{c}{GPT-4o}\\
  \hline
  Data & \multicolumn{2}{c}{Audio} & \multicolumn{2}{c|}{Video} & \multicolumn{2}{c}{Audio} & \multicolumn{2}{c}{Video}\\
  \hline
  Metric & SR & Cost & SR & Cost & SR & Cost & SR & Cost\\
  \hline
  -I & 0.96 & 0.15 & 0.95 & 0.19 & 0.91 & 0.15 & 0.85 & 0.17\\
  \hline
  -G & 0.70 & 0.08 & 0.69 & 0.12 & 0.50 & 0.11 & 0.68 & 0.09\\
  \hline
  -IG & 0.58 & 0.22 & 0.66 & 0.11 & 0.46 & 0.21 & 0.66 & 0.13\\
  \hline
  -LN & 0.84 & 0.10 & 0.82 & 0.03 & 0.79 & 0.08 & 0.77 & 0.12 \\
  \hline
  -AE & 0.65 & 0.08 & 0.80 & 0.07 & 0.72 & 0.06 & 0.83 & 0.09 \\
  \hline
  -SS & 0.04 & 0.11 & 0.05 & 0.03 & 0.05 & 0.08 & 0.04 & 0.11 \\
  \hline
  Origin & 0.95 & 0.09 & 0.93 & 0.05 & 0.90 & 0.11 & 0.88 & 0.10\\
  \hline
  \end{tabular}
  \vspace{-0.02in}
\end{table}

\subsection{Hyperparameter Study (RQ3)}

\begin{figure}[t]
  \centering
  \begin{subfigure}{0.23\textwidth}
    \centering
    \includegraphics[width=\textwidth]{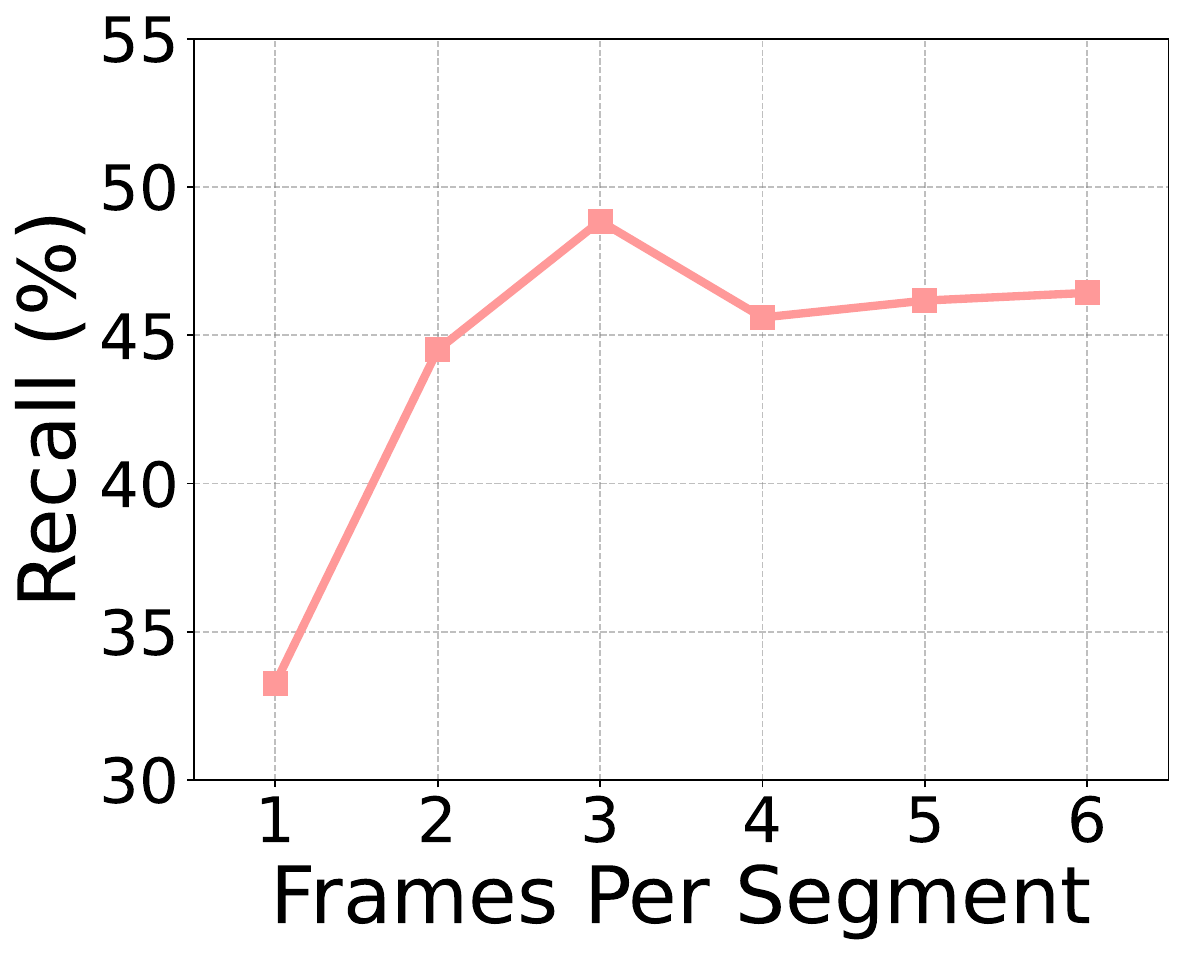}
    \caption{Recall Evaluation}
    \label{fig:recall_performance}
  \end{subfigure}
  \hfill
  \begin{subfigure}{0.23\textwidth}
    \centering
    \includegraphics[width=\textwidth]{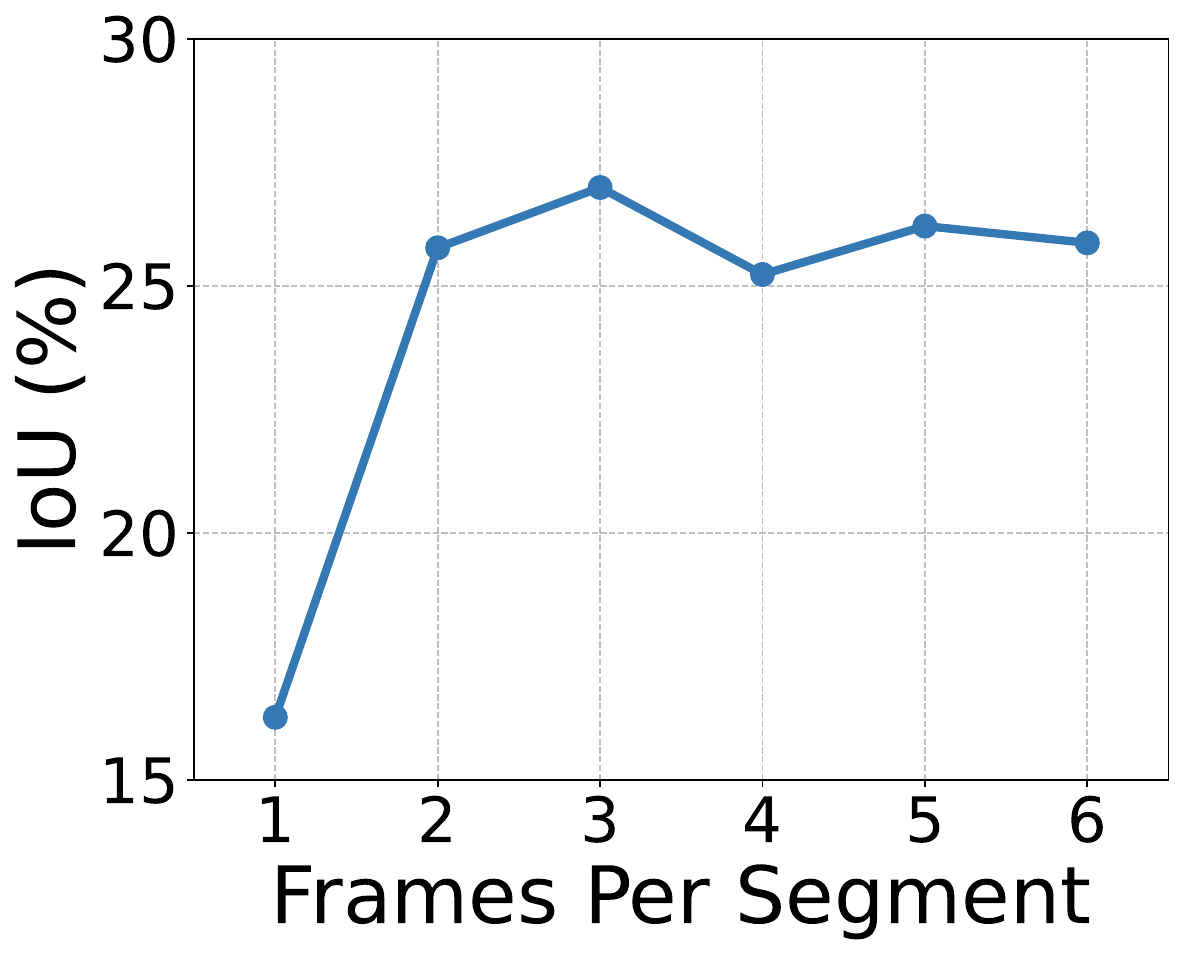}
    \caption{IoU Evaluation}
    \label{fig:iou_performance}
  \end{subfigure}
  \hfill
  \begin{subfigure}{0.23\textwidth}
    \centering
    \includegraphics[width=\textwidth]{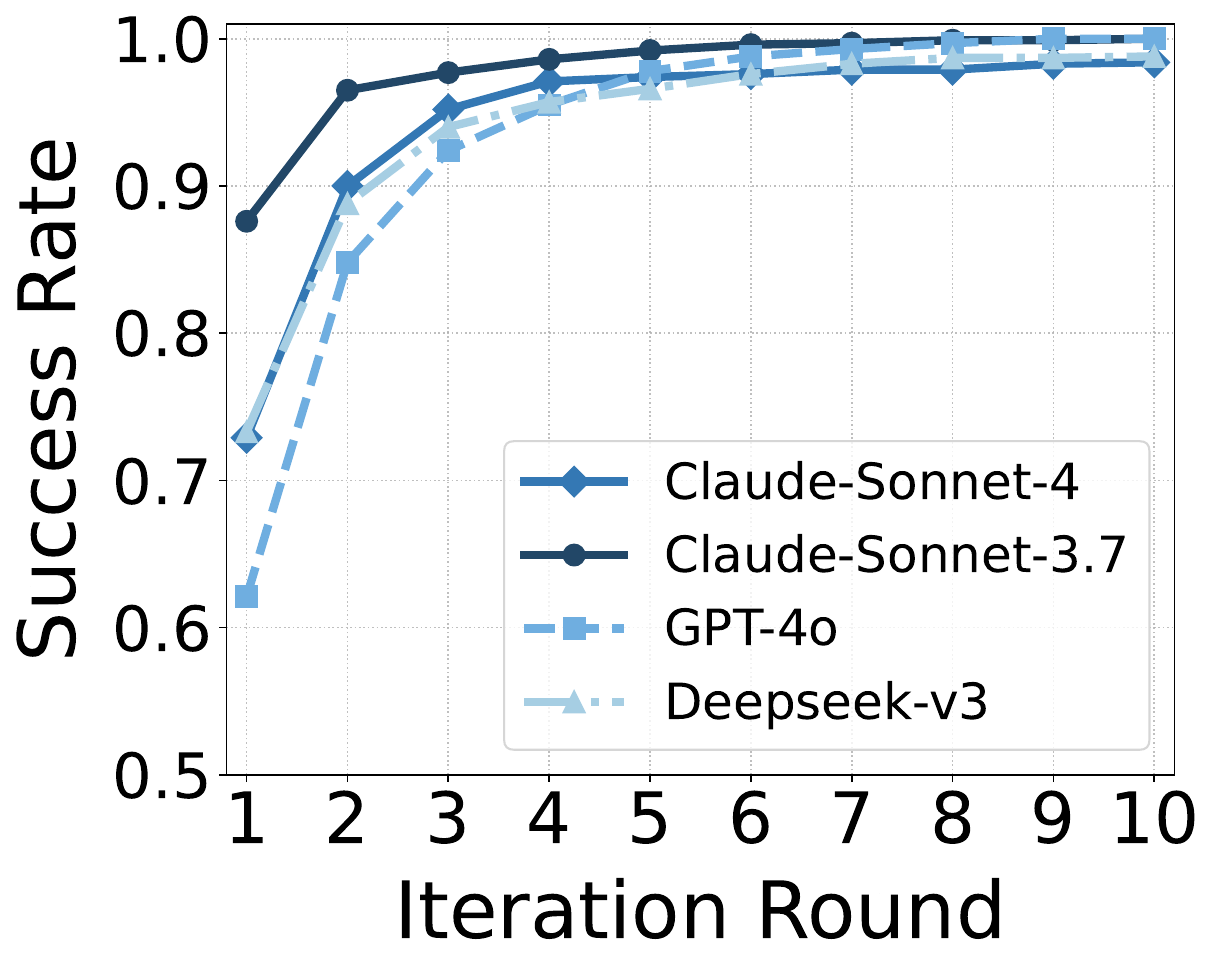}
    \caption{Audio Dataset}
    \label{fig:hyperaudio}
  \end{subfigure}
  \hfill
  \begin{subfigure}{0.23\textwidth}
    \centering
    \includegraphics[width=\textwidth]{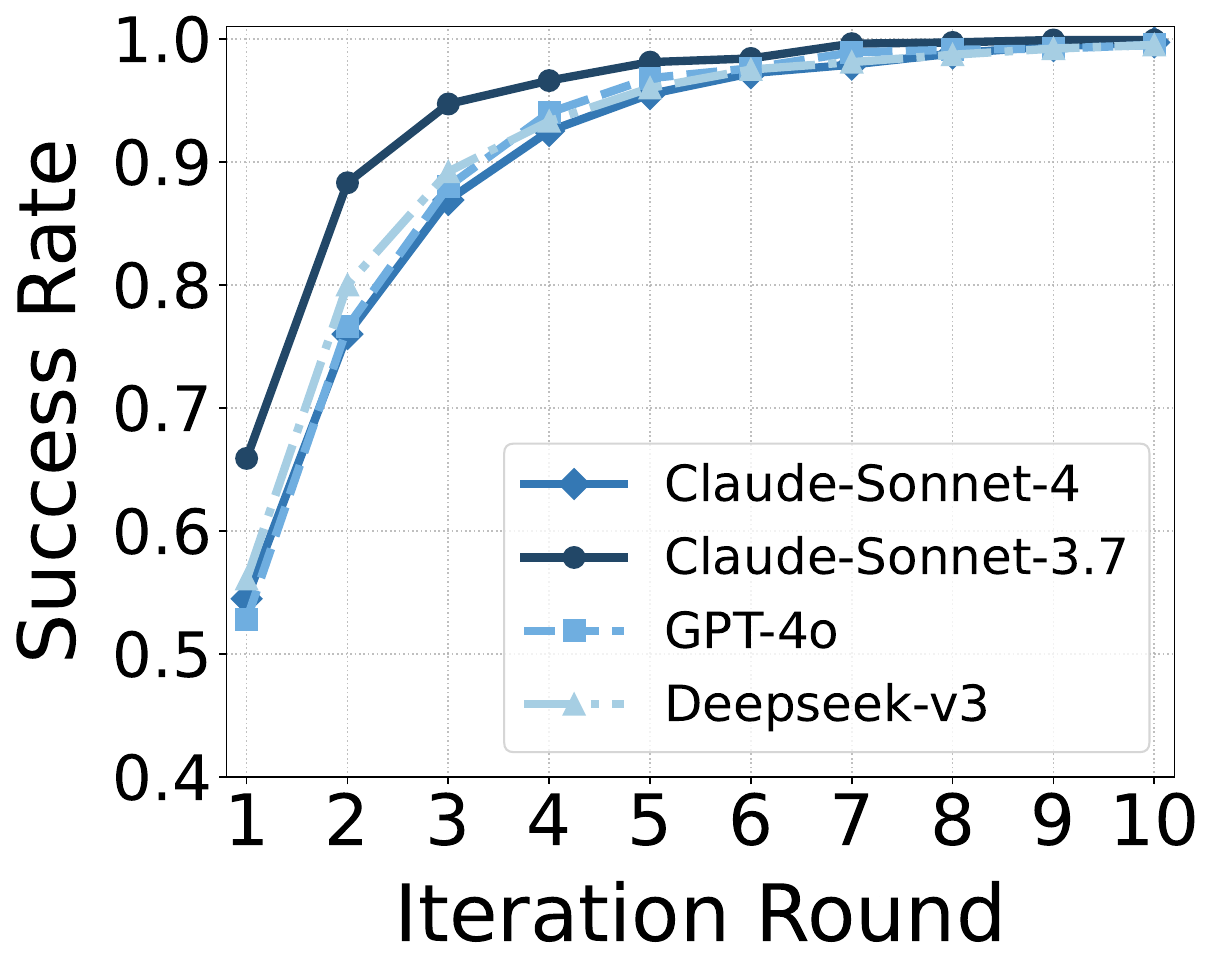}
    \caption{Video Dataset}
    \label{fig:hypervideo}
  \end{subfigure}
  \vspace{-0.05in}
  \caption{Hyperparameter study for \model.}
  \label{fig:hyperparameter_study}
  \vspace{-0.2in}
\end{figure}

This section studies the impact of two key hyperparameters in our \model. The evaluation results are shown in Figure~\ref{fig:hyperparameter_study}. From the results we make the following analysis:

\textbf{Frames per Segment} in shot planning affects the number of key frames extracted for each raw video resource $m_i$. Using only 1 frame results in a substantial decrease in recall performance, indicating insufficient visual information extraction capability. When the frame count is excessively increased, performance slightly declines, potentially due to over-sampling introducing noisy frames.

\textbf{Optimization Round} in workflow orchestration shows that increasing Optimization rounds consistently improves the success rate of workflow orchestration, demonstrating typical performance scaling during inference. The results also indicate that sufficient gradient-based graph iterations can significantly reduce performance gaps between different underlying LLM models.

\vspace{-0.05in}
\subsection{LLM-Human Judge Consistency (RQ4)}
\label{sec:consistency_study}
\begin{figure*}[t]
  \centering
  \includegraphics[width=1\columnwidth]{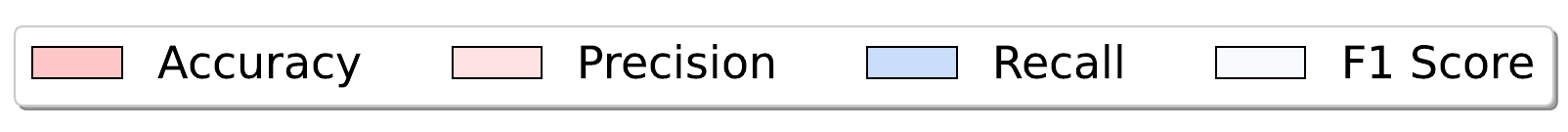}
  
  \begin{subfigure}[b]{0.64\columnwidth}
    \centering
    \includegraphics[width=\textwidth]{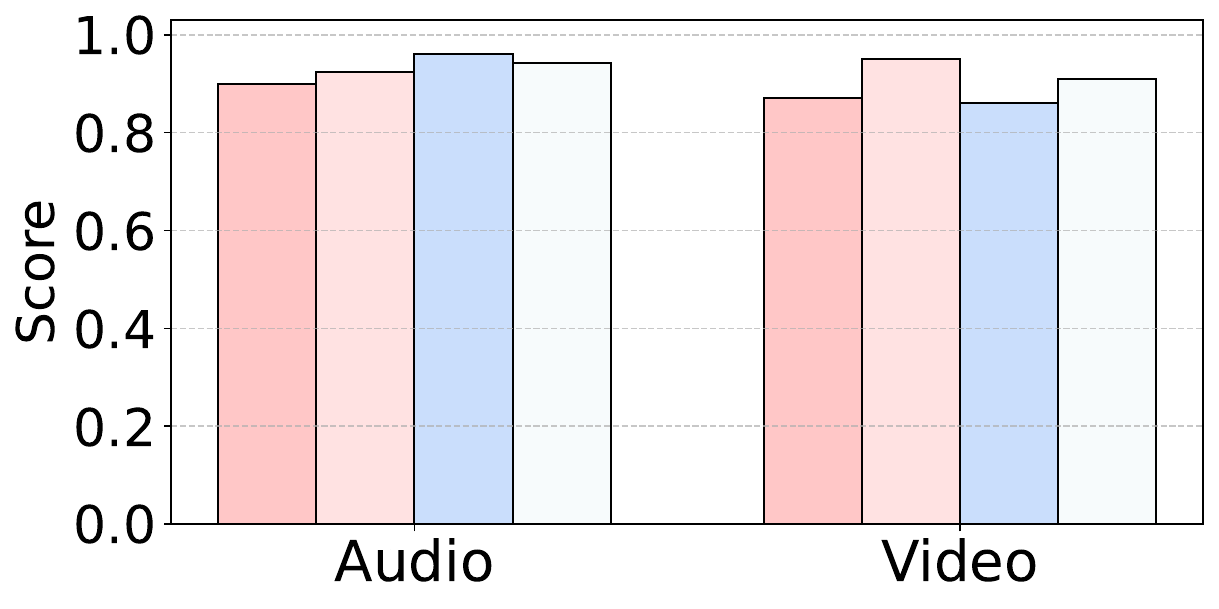}
    \label{fig:consistency_claude}
    \vspace{-0.26in}
    \caption{Claude-Sonnet-3.7}
  \end{subfigure}
  \hfill
  \begin{subfigure}[b]{0.64\columnwidth}
    \centering
    \includegraphics[width=\textwidth]{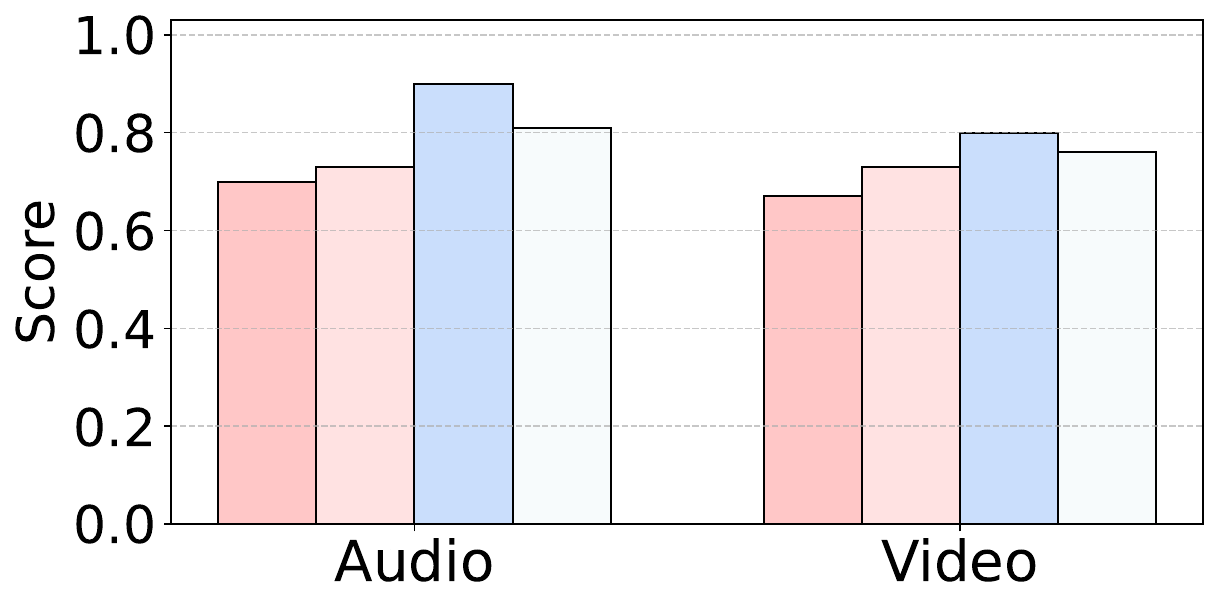}
    \label{fig:consistency_deepseek}
    \vspace{-0.26in}
    \caption{Deepseek-v3}
  \end{subfigure}
  \hfill
  \begin{subfigure}[b]{0.64\columnwidth}
    \centering
    \includegraphics[width=\textwidth]{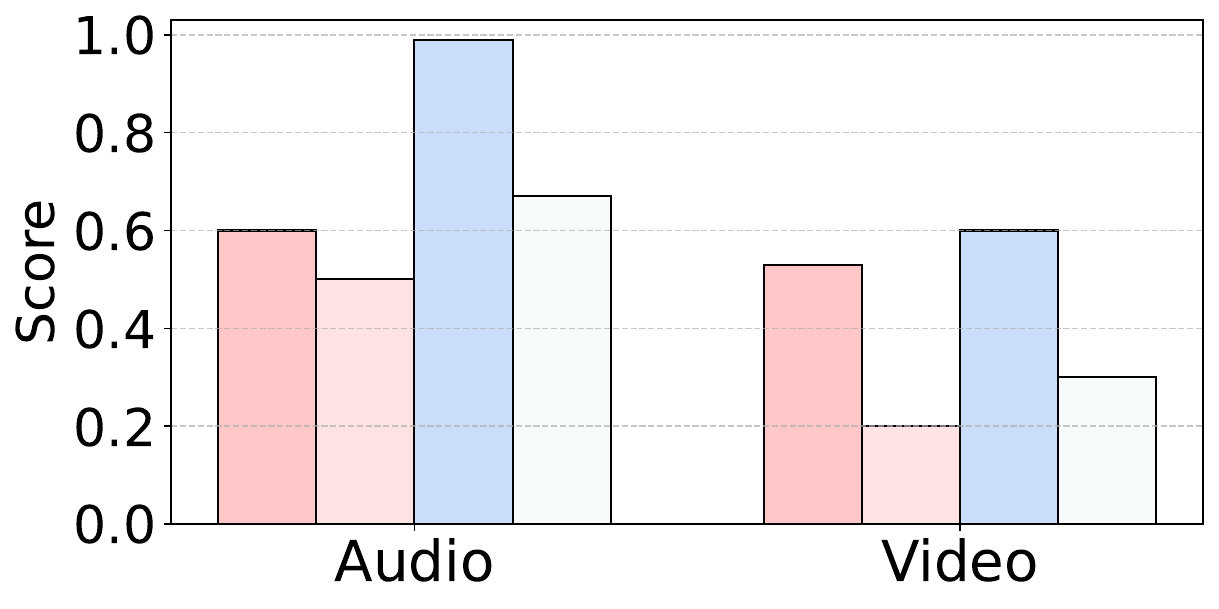}
    \label{fig:consistency_gpt}
    \vspace{-0.26in}
    \caption{GPT-4o}
  \end{subfigure}
  \vspace{-0.1in}
  \caption{Consistency study on LLM self-evaluation}
  \label{fig:consistency}
  \vspace{-0.15in}
\end{figure*}

We assess the reliability of LLM-based evaluation by measuring its agreement with human annotations. Specifically, we generate 30 agent graphs via \model\ across diverse creation instructions, manually label their success status, and compute accuracy, precision, recall, and F1 against human judgments. Results are shown in Figure~\ref{fig:consistency}. Claude-Sonnet-3.7 achieves the strongest alignment with human judgment, scoring 0.85--1.0 across all metrics. Across model-modality pairs, recall consistently exceeds precision, indicating that LLM evaluators are more prone to false positives than false negatives. This suggests our framework rarely overlooks genuinely successful orchestrations.

\begin{figure*}[t]
  \centering
  \includegraphics[width=\linewidth]{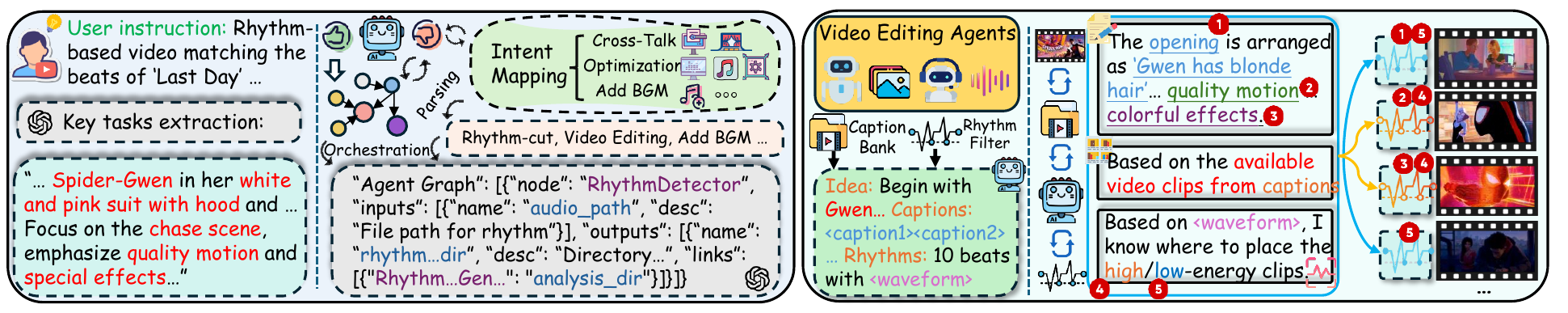}
  \vspace{-0.25in}
  \caption{Case study: Creating a rhythm-synced Spiderman movie montage with \model.} 
  \label{fig:casestudy}
  \vspace{-0.2in}
\end{figure*}

\subsection{Case Study on Created Videos (RQ5)}
We present a case study to illustrate \model's practical capabilities. As shown in Figure~\ref{fig:casestudy}, given an instruction to create a rhythm-synced \emph{Spiderman} montage, the system identifies the high-energy nature of the task and adapts its agent graph for fast-paced editing. It synchronizes visual cuts with musical beats and retrieves target scenes from shot-level descriptions, demonstrating seamless multimodal integration. This example highlights how \model\ dynamically composes agents based on user intent and available resources, transitioning from basic audio extraction to sophisticated rhythm-synced editing within a unified framework.

\vspace{-0.1in}
\section{Related Work}
\label{sec:relate}
\vspace{-0.1in}

\noindent\textbf{Automated Multimodal Editing and Generation} integrates multiple modalities for content creation and manipulation. Video methods enforce spatiotemporal consistency~\citep{wang2025videodirector}, while multimodal LLMs parse complex instructions~\citep{zhang2024moonshot, cheng2025text}. Recent generative advances~\citep{zhang2025deep, Mao2025OmniEffectsUA, Ren2024ConsistI2VEV} produce impressive outputs, yet these results still require substantial editing for real-world scenarios such as vlog creation and story reenactment. While prior work excels at individual editing capabilities, it lacks systematic orchestration of interdependent operations across complex multi-step tasks. We address this gap by integrating over 30 specialized agents with a dynamic graph-based orchestration framework that constructs and refines task-specific graphs.

\noindent\textbf{Multi-Agent Systems}. Recent multi-agent research moves from monolithic models to collaborative frameworks where multiple autonomous agents dynamically cooperate on complex tasks~\citep{zhuge2024gptswarm, tang2025autoagent}. Large language models play a central role through natural language reasoning and tool use, enabling dynamic role specialization in evaluation and refinement, program synthesis and execution~\citep{tang2025ai}, and interaction with external APIs, databases, and computational resources~\citep{shi2024learning}. Existing systems, however, are largely confined to text-centric applications such as coding, mathematics, and beyond. Our work extends this paradigm to multimodal video editing workflows.

\noindent\textbf{Video Understanding and Retrieval} are intrinsically connected pillars of visual intelligence, with advances in one domain driving progress in the other. Modern research in video understanding focuses on developing specialized models for challenges like visual question answering (VQA)~\citep{qian2024nuscenes, sima2024drivelm, li2025patch}, leveraging complex fusion mechanisms to extract spatial and temporal semantics. Representative works such as VideoMind~\citep{liu2025videomind} and Qwen-2.5-VL~\citep{bai2025qwen2} demonstrate strong capabilities in long-video scenarios, enabling precise temporal localization and spatio-temporal reasoning. Our work differs by designing a comprehensive framework with strong video understanding and retrieval capabilities specifically tailored for practical video editing workflows.

\vspace{-0.1in}
\section{Conclusion}
\label{sec:conclusion}
\vspace{-0.1in}
We presented \model, a comprehensive agentic framework that enables automated video creation across diverse genres and production workflows. Our approach addresses two fundamental challenges: coherent long-form video planning through global-aware shot creation, and multi-agent workflow orchestration via textual-gradient graph optimization. \model\ integrates over thirty specialized editing agents and substantially outperforms existing methods, achieving 87-95\% orchestration success rates while reducing MLLM API costs by 60\%. Human evaluation confirms that \model\ produces high-quality videos comparable to human creators across multiple categories. These results establish \model\ as an effective solution for bridging the gap between content creation demand and technical accessibility.
\section*{Limitations}

Our study serves as an initial exploration of all-in-one agentic video editing, and several aspects warrant further investigation. First, \model's retrieval-based shot creation inherits the quality constraints of available source footage. Integrating generative video models could complement this pipeline for scenarios where suitable material is unavailable. Second, \model\ has built a community of nearly 300 members, demonstrating its practical value. However, we acknowledge that as the community expands, automated video editing capabilities could potentially be misused to create misleading or manipulative content that spreads misinformation. We encourage future work to explore content moderation frameworks and responsible usage practices to mitigate these risks.

\section*{Ethical Considerations}

\model\ uses publicly available materials for academic research, including popular-video information from online platforms and publicly accessible film and television resources. Our data collection focuses on public content and visible aggregate signals rather than private user information, and we do not collect, infer, or release personal or sensitive data. The resources and outputs in this work are used only for research and evaluation. We respect platform terms, copyright, and content-usage restrictions, and do not intend to redistribute protected source media or support commercial deployment. Beyond these data-use and responsible-use considerations, we do not identify additional ethical concerns specific to this work.

\bibliography{ref}

\clearpage
\appendix
\def\model{VideoAgent}
\section{Appendix}
\label{tab:appendix}

\begin{figure*}[t]
  \centering
  \includegraphics[width=\linewidth]{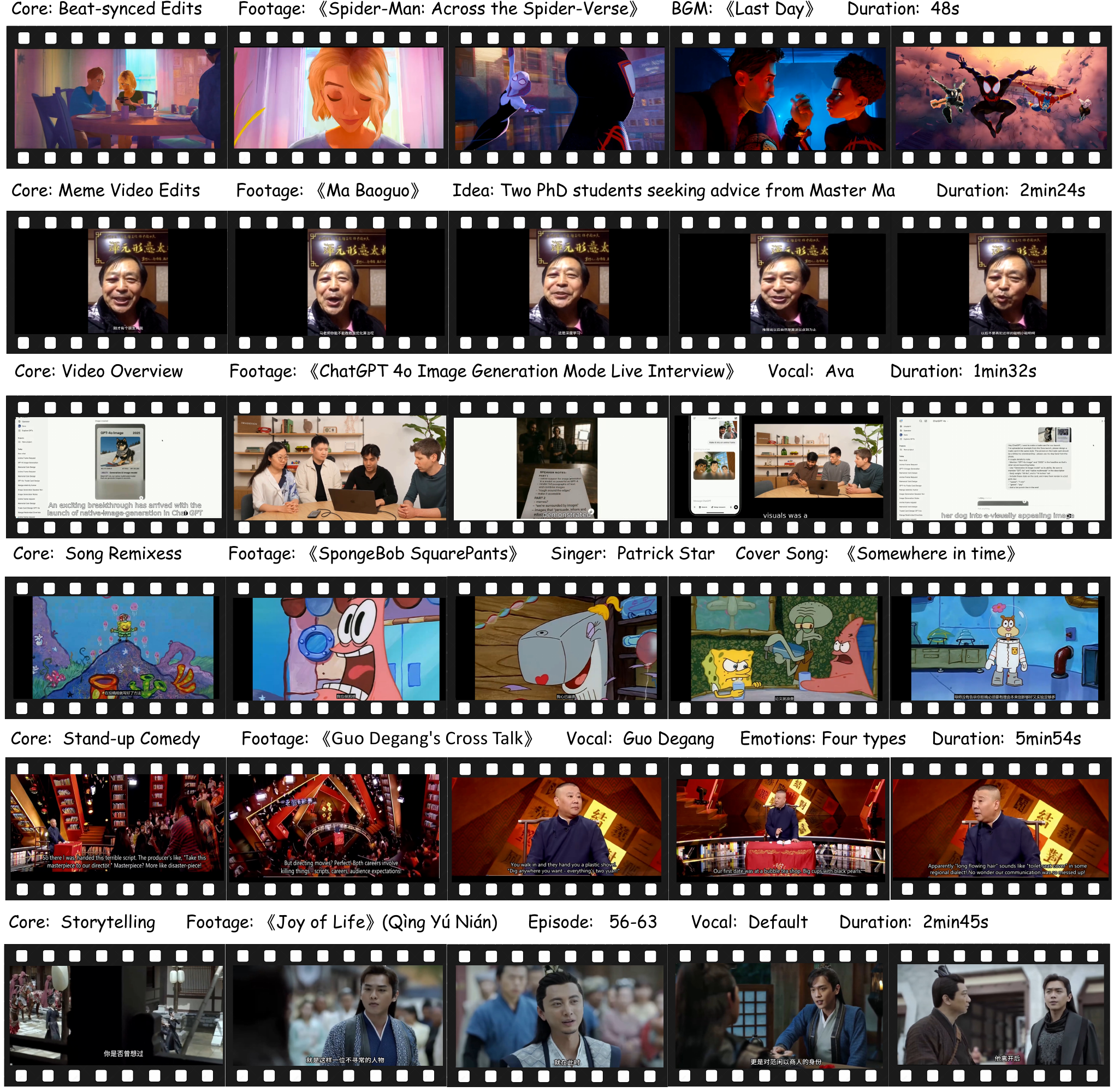}
  \caption{Video cases of \model\ in real-world scenarios - Case 1} 
  \label{fig:demo1}
\end{figure*}

\begin{figure*}[t]
  \centering
  \includegraphics[width=\linewidth]{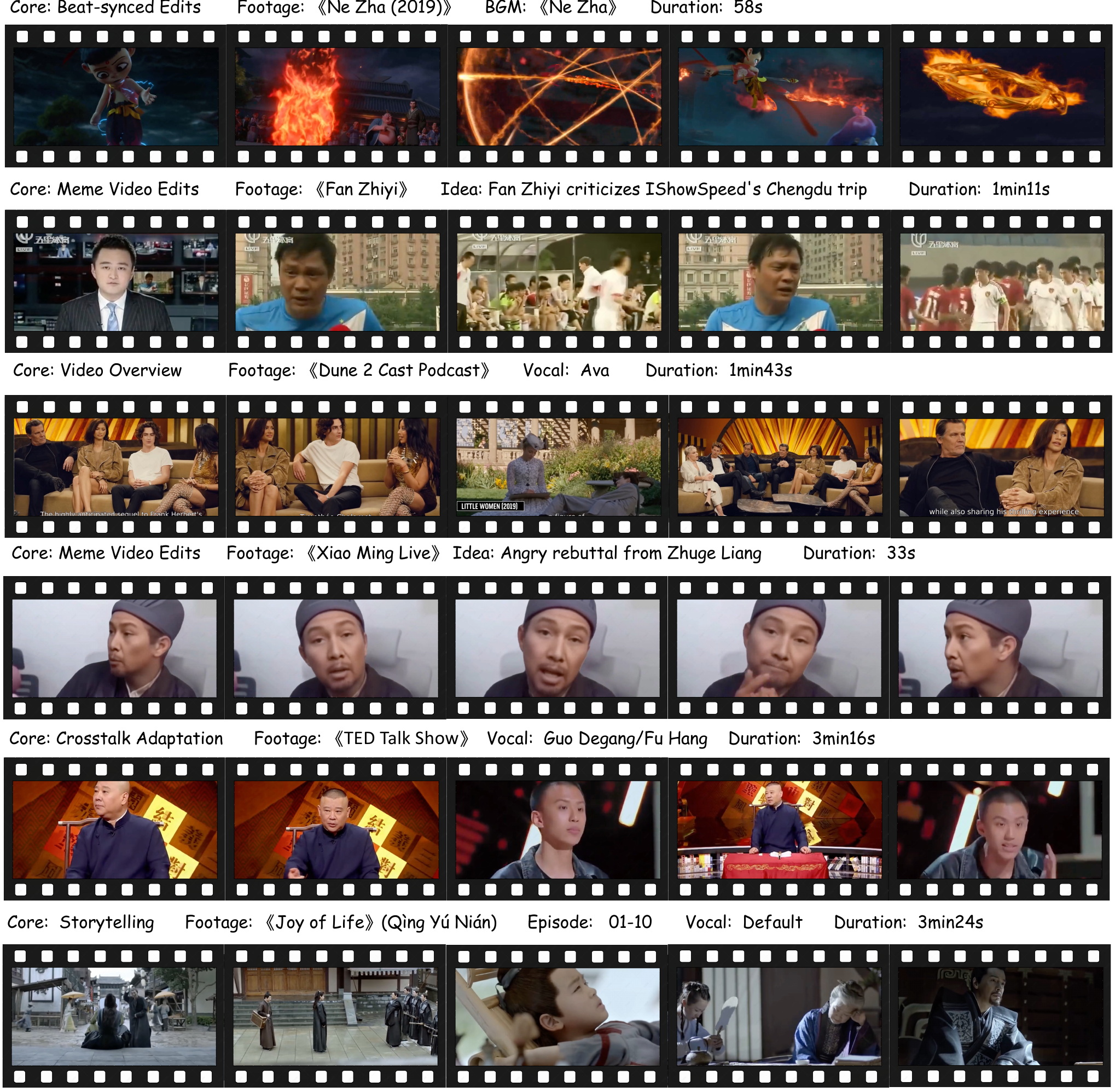}
  \caption{Video cases of \model\ in real-world scenarios - Case 2} 
  \label{fig:demo2}
\end{figure*}

\begin{figure*}[t]
  \centering
  \includegraphics[width=\linewidth]{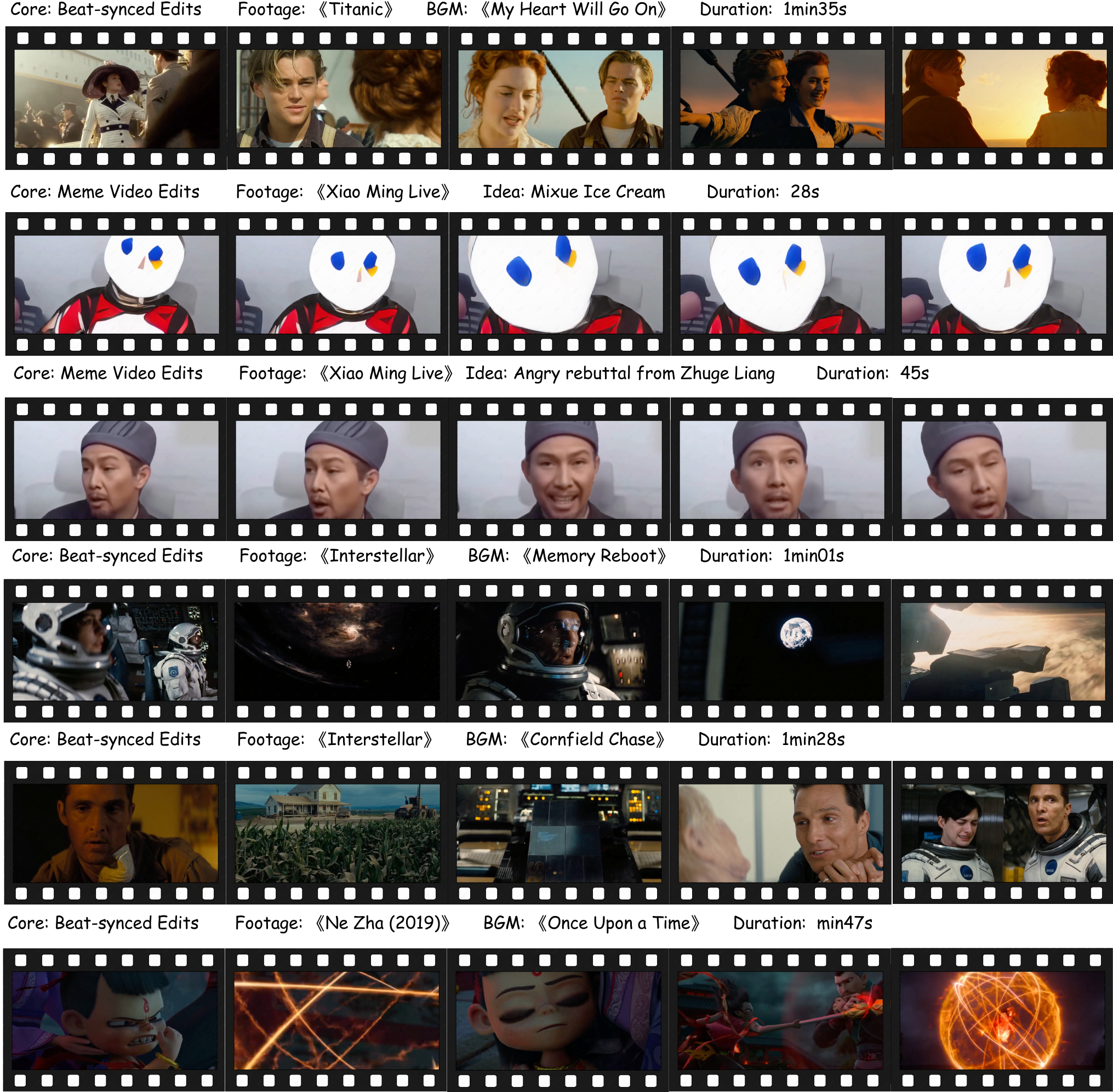}
  \caption{Video cases of \model\ in real-world scenarios - Case 3} 
  \label{fig:demo3}
\end{figure*}

\subsection{Video Showcase}\label{sec:video-showcase}

Below are video showcases we created with VideoAgent, which can be grouped into six main categories: \textbf{Beat-synced Edits}, \textbf{Video Overviews}, \textbf{Storytelling}, \textbf{Meme Video Edits}, \textbf{Song Remixes}, and \textbf{Cross-lingual Adaptations} (including stand-up comedy and crosstalk adaptations). 

\textbf{Beat-synced Edits}: The user provides background music and film footage. VideoAgent automatically detects beat and tempo changes in the music and aligns high-energy film visual shots with strong beat moments. \textbf{Video Overview}: The user provides video footage of a news item or event. VideoAgent transcribes the speech, writes a summary of the event, matches the meaning of each summary sentence to corresponding visual segments from the original video, and generates voiceover for each sentence to complete the edit. \textbf{Storytelling}: The user provides plain text of a novel or story along with video footage for visual matching. VideoAgent scripts the story in a specific narration style, matches each line of the script to suitable video segments based on semantic meaning, and generates voiceover for each script segment to produce the final video. \textbf{Meme Video Edits}: The user supplies a source video and a custom script or narrative. VideoAgent extracts and transcribes the original audio, generates new speech, precisely syncs the replacement audio to video frames, and outputs a professionally dubbed, meme-style edit. \textbf{Song Remixes}: The user provides a MIDI file, lyrics, background music, and a target voice sample. VideoAgent generates a cover, clones the target voice, aligns timing with the arrangement, and integrates the result into the video pipeline for a polished, synced remix. \textbf{Cross-lingual Adaptations}: The user provides source audio and target voice samples. VideoAgent adapts the script into the desired cultural format (e.g., Chinese crosstalk or English talk show), synthesizes target voices, applies appropriate effects, and integrates the result into the video editing pipeline for a polished, synced adaptation.

Detailed prompts and examples for each showcase category are presented in Listing~\ref{lst:videoagent-showcase1}, Listing~\ref{lst:videoagent-showcase2}, and Listing~\ref{lst:videoagent-showcase3}.

\subsection{Evaluation Datasets and Protocols}\label{sec:data_protocol}

\begin{figure}[htbp]
  \centering
  \includegraphics[width=\columnwidth]{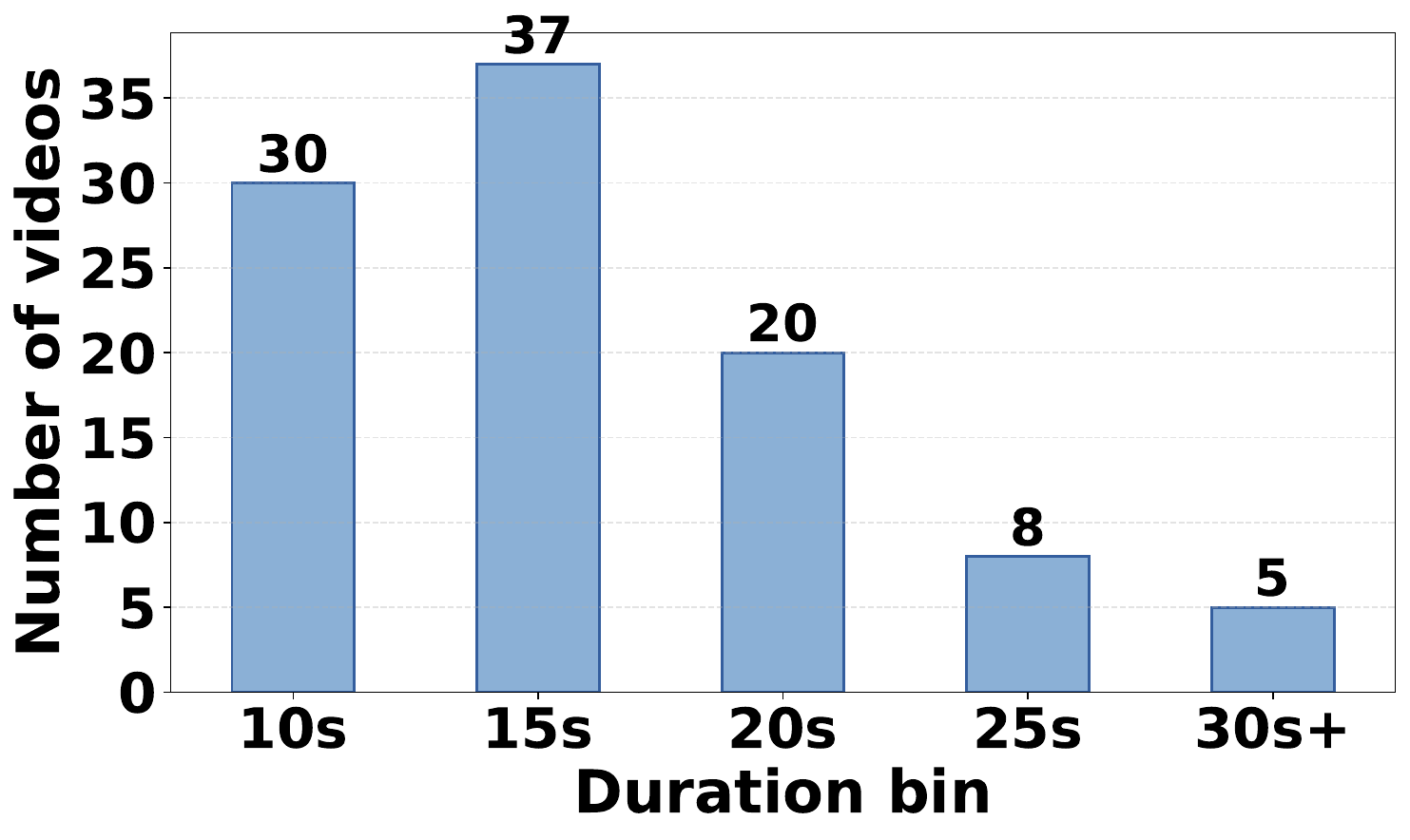}
  \caption{Evaluation video durations distribution.}
  \label{fig:video_duration}
\end{figure}

\begin{figure}[htbp]
  \centering
  \includegraphics[width=\columnwidth]{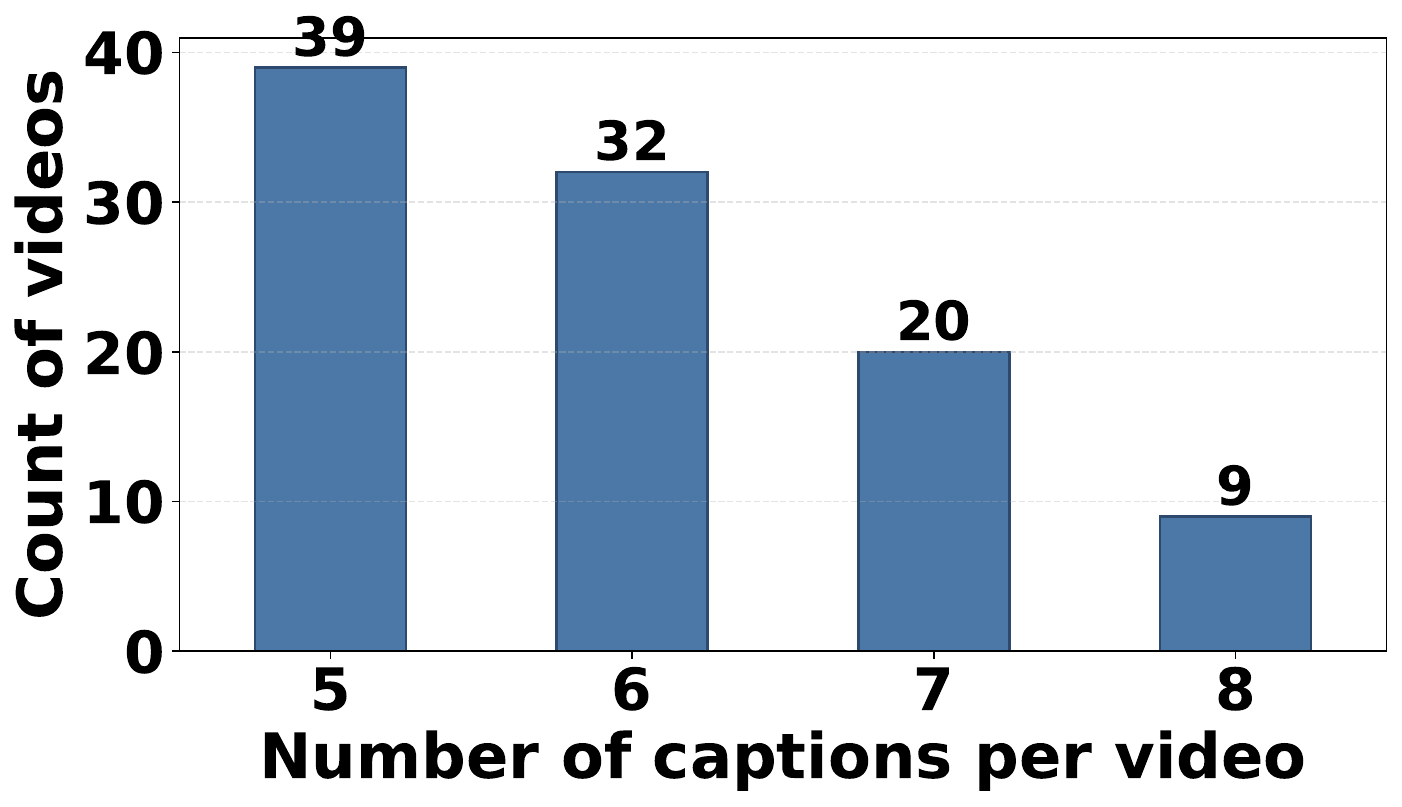}
  \caption{Evaluation video captions/queries distribution.}
  \label{fig:captions_count}
\end{figure}

To evaluate \model\ in terms of both video understanding and workflow orchestration, we conduct experiments from the following two perspectives.
\begin{itemize}[leftmargin=*]
\item \textbf{Video Understanding and Retrieval Evaluation}. We utilize the \textbf{Shot2Story} dataset~\cite{han2023shot2story20k} to evaluate whether \model\ can effectively retrieve appropriate video clips according to user inputs. This dataset contains video clips paired with corresponding queries that describe each clip's content. We randomly select 100 videos with more than 5 queries each, using the midpoint frames as keyframes for retrieval evaluation. In total, there are 599 individual queries. Note that achieving such global understanding and precise retrieval across multiple video segments is computationally intensive. Processing all queries for a multimodal LLM or agent typically requires around {3-9} hours. The distribution of evaluation video durations in our sampled dataset is shown in Figure ~\ref{fig:video_duration}. Videos are grouped into fixed intervals 10--15s, 15--20s, 20--25s, 25--30s, and 30s+ based on their total segment lengths. The caption queries per video distribution in our dataset is illustrated in Figure ~\ref{fig:captions_count}. Each video in the dataset contains between 5 and 8 captions, providing multiple query perspectives for evaluation. The dataset exhibits a diverse range of content categories, with Entertainment, Film \& Animation, Autos \& Vehicles, Sports , News \& Politics, Howto \& Style, Science \& Technology, Comedy, Education, Travel \& Events, People \& Blogs, Gaming and Pets \& Animals of the sampled videos. \vspace{0.02in} 

    We use three metrics to compare retrieved video clips and groundtruth shots: \textbf{i) Recall:} This is the percentage of groundtruth videos being retrieved with the first rank for each query. \textbf{ii) Embedding Matching Score:} Following~\cite{cheng2025text}, we compute the coarse-grained embedding matching score between retrieved video content and LLM-generated captions (see List ~\ref{lst:video_caption_transform_prompt}) using ImageBind~\cite{Girdhar2023ImageBindOE} cross-modal encoder. This evaluates whether retrieved videos align with user intents at a semantic level. \textbf{iii) Intersection over Union:} We compute temporal IoU between predicted and ground-truth video segments. IoU measures the overlap between retrieved and target clips, ranging from 0 to 1.

    \item \textbf{Workflow Orchestration Evaluation}. To assess the model's performance in orchestrating workflows for video editing instructions, we construct the \textbf{VideoEdit} dataset containing 2000 creation instructions encompassing diverse visual and audio editing requirements. Starting with authentic video editing demands, such as creating prompts for ChatGPT-4o image generation live stream overviews like "Short movie podcast, colloquial expression within 300 words, notice to identify which actor or host is talking, don't mention movie tickets available issue", details of prompts can see Appendix~\ref{sec:video-showcase}. We then use Claude-Sonnet-3.7 for data augmentation to generate additional creative requirements, the audio editing and video editing system prompts are presented in List~\ref{lst:audio_editing_instruction_prompt} and List~\ref{lst:video_editing_instruction_prompt}. For audio editing, examples include requirements such as "I want to create a funny stand-up comedy audio with audience reactions. The content should be adapted from a script I provide, and I'd like the comedian's voice to match a specific target vocal style." For video editing, examples include "I have a video that I need to edit with a different script while maintaining the original speaker's voice. I want to keep the visuals exactly the same, but change what the person is saying to better match my needs. The new content should sound natural and match the original speaker's voice and speaking style." Through this approach, we yield 2000 distinct instructions covering the full spectrum of video editing tasks.\vspace{0.02in}

    We employ Claude-Sonnet-3.7 to build a \textbf{Judge Agent} that evaluates whether \model\ and baseline models successfully orchestrate agent workflows for these complex requirements. This judge agent verifies instruction fulfillment and validates data flow compatibility across agents, ensuring outputs from preceding agents are properly formatted as inputs for subsequent agents. The judge agent demonstrates high alignment with human evaluators in comparative testing (see Section~\ref{sec:consistency_study}). This evaluation uses \textbf{Success Rate} as the metric, indicating the percentage of samples where the orchestrated workflow fulfills the user instruction.
\end{itemize}

\subsection{Baseline Method}\label{sec:baseline}
To ensure the comprehensiveness of the benchmark, we conducted a multi-dimensional comparison with n baselines encompassing different mechanisms.\\  
\textbf{Agentic Systems on Different Domains.}  
\setlength{\leftmargini}{10pt}  
\begin{itemize}  
    \item {\textbf{Chain-of-Thought}} \cite{wei2022chain}:This method proposes a chain of thought to improve LLMs' ability to perform complex reasoning by integrating explicit, structured intermediate steps.
    \item {\textbf{LLM Debate}} \cite{du2023improving}: It thoughtfully and systematically integrates debate results from multiple independent language model instances across several iterative rounds of refinement.
    \item {\textbf{Step Back}} \cite{zheng2023take}: This method enables LLMs to abstract from instances containing specific details to generate high-level concepts and foundational first principles efficiently. 
    \item {\textbf{ExpertPrompting}} \cite{xu2023expertprompting}: It leverages the potential of LLMs by delegating complex tasks to domain-specific expert agents and automatically synthesizing their respective opinions.  
    \item {\textbf{Intelligent Go-Explore}} \cite{lu2024intelligent}: This method is designed to replace handcrafted heuristics with intelligence, iteratively returning to observed states and searching for optimal solutions.
    \item {\textbf{Flow}} \cite{niu2025flow}: This framework introduces modular, parallel workflows to boost efficiency and adaptability, enabling real-time updates that preserve global coherence during challenges.
    \item {\textbf{GPTSwarm}} \cite{zhuge2024gptswarm}: This framework uses graphs to unify LLM agents with reinforcement learning, automatically optimizing prompts and agent orchestration for improved efficiency and adaptability. 
    \item {\textbf{GraphCounselor}} \cite{gao2025graph}:
    This method uses multi-agent collaboration with adaptive graph information extraction to dynamically adjust reasoning depth and improve semantic consistency in specialized domains.
\end{itemize}
\textbf{MLLM and Agents.}  
\setlength{\leftmargini}{10pt}
\begin{itemize} 
    \item {\textbf{Qwen2.5-VL}} \cite{bai2025qwen2}: This vision-language model is designed for long-video comprehension spanning hours, with precise second-level event localization and multimodal reasoning.
    \item {\textbf{VideoRAG}} \cite{ren2025videorag}: It introduces a RAG framework for long-context video understanding with graph-grounded and multimodal encoding, enabling unlimited retrieval and beating SOTA.
    \item {\textbf{VideoMind}} \cite{liu2025videomind}: This method offers a role-based workflow for temporally grounded video understanding, using Chain-of-LoRA for efficient role-switching via lightweight adapters.
    \item {\textbf{TeaserGen}} \cite{TeaserGen}: This framework focuses specifically on documentary teaser generation using pre-trained highlight detection modules with local learning-to-retrieval modules.
\end{itemize}
\textbf{Video Generation Agents.}  
\setlength{\leftmargini}{10pt}
\begin{itemize} 
    \item {\textbf{NoteBookLM}} \cite{notebooklm}: NotebookLM is a personalized AI knowledge assistant that leverages advanced LLMs to analyze and interact with user-provided documents collaboratively.
    \item {\textbf{Director}} \cite{director}: This framework is capable of reasoning through video-related tasks while simultaneously streaming the processed results in real-time to facilitate video synthesis.
    \item {\textbf{FunClip}} \cite{funclip}: FunClip is an open-source tool for automated video clipping. Users can quickly extract video segments by selecting recognized text or speaker segments.
    \item {\textbf{NarratoAI}} \cite{narratoaI}: NarratoAI is a video narration platform, combining script writing, editing, voice-over generation, and subtitles into a unified workflow for content production.
\end{itemize}

\subsection{Implementation Details}
\label{sec:implement}
This section will provide detailed descriptions of the implementation details and parameter configurations for each baseline in different experiments.

All baseline methods are configured using their default or recommended parameter settings and implemented following the original papers. All experiments were conducted on NVIDIA RTX 3090 GPUs. \textbf{Workflow Orchestration Evaluation}: In the workflow orchestration performance comparison and workflow orchestration ablation study, for fairness, we use Claude-Sonnet-3.7 as the Judge Agent to evaluate Agent Graphs and Agent Chains from multiple perspectives, including execution sequence validation, parameter routing correctness, agent functional redundancy assessment, and requirement fulfillment analysis. We also set the number of reflection rounds to 2 for all self-reflective methods. In the LLM-human judgment consistency experiment, to reduce the error rate of human judgment, we first execute agent graph checks for execution sequence validation and parameter routing correctness issues, then proceed with agent functional redundancy assessment and requirement fulfillment analysis. \textbf{Video Editing Evaluation}: All video inputs are segmented into three-second segments, under fps1, one frame is sampled per second. For recall calculations, the midpoint timestamp of each retrieved video segments serves as the reference point. Subsequently, two-second clips are extracted, one second before and one second after the midpoint timestamp, yielding a total duration of two seconds per retrieved clip. These extracted clips are then concatenated sequentially to generate the final output video for computing the Embedding Matching score and Intersection over Union metrics, all ours VLM agent/caption use MiniCPM-V 2.6 int4. In the video editing ablation study and hyperparameter experiments, all our base models utilize GPT-4o, and continue to keep the video segments divided into three seconds for input.

\subsection{VideoRepurpose Benchmark Evaluation}
\label{sec:videorepurpose_eval}

To further validate the effectiveness and superiority of our proposed framework, we extended our evaluation to include the VideoRepurpose benchmark~\cite{Wu2024VideoRF}. VideoRepurpose is a large-scale dataset designed for video repurposing from user-generated content, providing a comprehensive testbed for shot retrieval tasks. We randomly sampled 50 videos from the VideoRepurpose dataset, each approximately 1 minute long and containing 1-2 highlight time intervals. The evaluation was conducted following the standard protocol with Recall and Intersection over Union (IoU) as the primary metrics.

Table~\ref{tab:videorepurpose_results} presents the experimental results comparing our method against several strong baselines, including both commercial models and specialized video understanding models and frameworks. Our VideoAgent framework consistently achieves superior performance across both metrics. These results indicate that VideoAgent achieves the best performance on the VideoRepurpose benchmark, thereby further validating the effectiveness and superiority of the proposed framework across diverse datasets and evaluation scenarios.

\begin{table*}[h]
\centering
\caption{Performance comparison on VideoRepurpose benchmark.}
\label{tab:videorepurpose_results}
\begin{tabular}{cccccc}
\hline
Method & Recall & IoU & Method & Recall & IoU \\
\hline
GPT-4o & 41.27 & 19.31 & TeaserGen & 18.46 & 16.41 \\
Gemini-2.5-pro & 43.12 & 20.68 & Ours-GPT-4o & 47.13 & 19.75 \\
Claude-3.7-sonnet & 42.19 & 17.64 & Ours-Gemini-2.5-pro & 43.18 & 19.68 \\
VideoMind-7B & 40.11 & 16.89 & Ours-Claude-3.7-sonnet & 44.64 & 19.58 \\
\hline
\end{tabular}

\end{table*}

\subsection{Workflow Orchestration Efficiency Analysis}
\label{sec:workflow_efficiency}

To comprehensively evaluate the computational efficiency of VideoAgent, we conducted detailed latency measurements across different workflow orchestration strategies and backbone models. Table~\ref{tab:workflow_latency} presents the system latency for workflow orchestration in seconds, comparing four different orchestration approaches across various backbone LLMs, with separate measurements for audio and video processing tasks. The experimental results demonstrate that VideoAgent achieves a balanced trade-off between effectiveness and efficiency. Across different backbone models, VideoAgent consistently maintains competitive latency while delivering superior task completion quality. 
This balanced approach makes VideoAgent particularly suitable for real-world applications where both efficiency and effectiveness are critical. The results validate that VideoAgent's multi-agent orchestration framework can efficiently coordinate complex video processing workflows while maintaining computational efficiency across diverse backbone models and task types.

\begin{table*}[h]
\centering
\caption{System latency of workflow orchestration (in seconds) across different backbone models and orchestration strategies.}
\label{tab:workflow_latency}
\begin{tabular}{cccccccccc}
\hline
\multirow{2}{*}{Backbone} & \multicolumn{2}{c}{Claude-4} & \multicolumn{2}{c}{Claude-3.7} & \multicolumn{2}{c}{GPT-4o} & \multicolumn{2}{c}{Deepseek-v3} \\
\cmidrule(lr){2-3} \cmidrule(lr){4-5} \cmidrule(lr){6-7} \cmidrule(lr){8-9}
& Audio & Video & Audio & Video & Audio & Video & Audio & Video \\
\hline
Data Flow & 112.31 & 91.76 & 103.75 & 89.36 & 98.47 & 102.88 & 109.18 & 115.17 \\
GPTSwarm & 79.63 & 101.32 & 76.84 & 97.45 & 43.29 & 51.57 & 76.87 & 60.32 \\
GraphCounselor & 24.78 & 38.43 & 26.31 & 33.85 & 37.02 & 30.85 & 45.08 & 21.46 \\
VideoAgent & 38.14 & 56.73 & 39.90 & 53.42 & 23.21 & 39.45 & 50.48 & 36.67 \\
\hline
\end{tabular}

\end{table*}

\subsection{Human Evaluation Details}
\label{sec:human evaluation}
To ensure fairness and transparency in human evaluation, we will provide detailed information regarding the experimental setup. We produced 19 demos using VideoAgent, and together with the baseline versions, there are 49 demos in total. They were organized into 18 pairs and evaluated by a total of 26 participants. participants voluntarily completed the anonymized evaluation and were informed that their ratings would be used for research evaluation. These participants rated each demo based on three criteria: consistency, audio quality, and scene diversity. The demos were uploaded to a shared drive and organized into a Google Form, with the order randomized and the files anonymized. This evaluation was conducted within a community group, which includes over 200 members from diverse backgrounds such as researchers, content creators, and hobbyists. Collecting all the ratings took about one week, and each participant spent approximately 1 hour to complete the questionnaire. 

\begin{table*}[h]
\centering
\caption{Human Evaluation across different methods}
\label{tab:human evaluation2}
\begin{tabular}{ccccccc}
\hline
Method & FunClip & Director & NotebookLM & NarratoAI & Handiwork & \model \\
\hline
Consistency & 1.9 & 2.0 & 1.5 & 1.1 & 3.5 & 3.2 \\
Scene diversity & 1.6 & 1.8 & - & 1.3 & 4.2 & 3.8 \\
Audio quality & 1.4 & 1.2 & 2.3 & 2.8 & 3.9 & 4.1 \\
Total Score & 1.6 & 1.7 & 1.9 & 1.7 & 3.9 & 3.7 \\
\hline
\end{tabular}
\end{table*}

\subsection{Tooluse Agents}\label{sec:tooluse}
To accommodate diverse user needs, we have designed a variety of multimodal tool-use agents capable of handling tasks such as audio preprocessing, video overview, storytelling, beat-synced edits, meme video remaking, song remixes, cross-lingual adaptations as well as video edits. The agent names and detailed descriptions of these tools are presented in Table \ref{tab:agentinfo}.

\subsection{Visual Processing Agents}\label{sec:visual}

Visual processing agents are specialized components that handle video content analysis, scene caption, and visual element manipulation for multimedia production workflows. These agents bridge the gap between textual content and visual representation by converting visual scenes into actionable narrative descriptions, managing video file operations, and coordinating the integration of visual and audio elements.

\subsubsection{VideoPreloader}

The VideoPreloader serves as the foundational initialization agent for video processing pipelines, establishing the necessary directory structure and preprocessing operations required for video editing workflows. It integrates with VideoRAG systems to enable efficient video content indexing and retrieval. The detailed algorithm structure is presented in Listing~\ref{lst:videopreloader}.

\subsubsection{VideoSearcher}

The VideoSearcher retrieves matching video clips from preprocessed video databases based on scene semantic descriptions. It leverages VideoRAG content indexing to perform intelligent video segment matching for automated video assembly workflows. The detailed algorithm structure is presented in Listing~\ref{lst:videosearcher}.

\subsubsection{VideoEditor}

The VideoEditor performs final video assembly by integrating matched video segments with synchronized audio tracks. It utilizes multimodal analysis to optimize clip selection and ensures temporal alignment between visual content and audio/storyboards narratives. The detailed algorithm structure is presented in Listing~\ref{lst:videoeditor}.

\subsubsection{FaceSwapping}
The FaceSwapping agent is capable of replacing a specified face in a video with that of a target person. It uses the Viggle AI tool to achieve precise face replacement, enhancing the overall viewing experience of the video. The detailed algorithm structure is presented in Listing~\ref{lst:faceswapping}.

\subsubsection{LipSynchronization}
The LipSynchronization agent is capable of synchronizing the speaker's lip movements in a video with a target audio track. It uses the Kling AI tool to achieve precise lip-sync alignment, enhancing the realism and coherence of the video's audio-visual experience. The detailed algorithm structure is presented in Listing~\ref{lst:lipsynchronization}.

\subsection{Audio Processing Agents}\label{sec:audio}

Audio processing agents constitute a fundamental component of the multimedia production pipeline, handling diverse audio manipulation tasks. These agents leverage advanced signal processing techniques and machine learning models to ensure high-quality audio output while maintaining compatibility across different formats and sampling rates. The audio processing workflow encompasses both basic preprocessing operations and sophisticated content analysis capabilities, with agents designed to handle everything from simple format conversions to complex multi-track audio synthesis.

\subsubsection{AudioExtractor}

The AudioExtractor agent provides comprehensive audio extraction capabilities from video sources, supporting both single file and batch directory processing. It utilizes FFmpeg for high-quality audio conversion while maintaining consistent output format specifications across all processed files. The detailed algorithm structure is presented in Listing~\ref{lst:audioextractor}.

\subsubsection{RhythmDetector}

The RhythmDetector agent provides comprehensive music rhythm analysis for rhythm-cut video creation workflows. It employs advanced signal processing techniques using librosa and scipy to detect beat patterns, analyze temporal distributions, and generate precise cut points for video editing synchronization with musical rhythms. The detailed algorithm structure is presented in Listing~\ref{lst:rhythmdetector}.

\subsubsection{LoudnessNormalizer}

The LoudnessNormalizer agent provides comprehensive audio loudness normalization capabilities for maintaining consistent audio levels across multiple files. It serves as a critical preprocessing component in audio production workflows, utilizing the FAP (Fast Audio Processing) tool for batch loudness standardization. The detailed algorithm structure is presented in Listing~\ref{lst:loudnessnormalizer}.

\subsubsection{Merge}

The Merge agent provides direct video and audio track combination capabilities without intermediate video clip processing. It distinguishes itself from other multimedia agents by performing complete file merging rather than segmented video editing, utilizing FFmpeg for high-quality encoding and synchronization. The detailed algorithm structure is presented in Listing~\ref{lst:merge}.

\subsubsection{Mixer}

The Mixer agent provides sophisticated audio mixing capabilities for combining foreground audio with background music tracks. It utilizes PyDub for comprehensive audio processing, supporting dynamic volume adjustment, length synchronization, and high-quality audio overlay operations. The detailed algorithm structure is presented in Listing~\ref{lst:mixer}.

\subsubsection{Resampler}

The Resampler agent provides comprehensive audio resampling capabilities for standardizing sample rates across multiple audio files within a directory. It leverages the FAP (Fast Audio Processing) framework to perform batch resampling operations with real-time progress monitoring and threaded output handling. The detailed algorithm structure is presented in Listing~\ref{lst:resampler}.

\subsubsection{Separator}

The Separator agent provides advanced audio source separation capabilities for isolating vocal and instrumental components from mixed audio tracks. It utilizes machine learning-based separation algorithms through the FAP framework to perform high-quality vocal extraction and background music isolation across multiple audio files. The detailed algorithm structure is presented in Listing~\ref{lst:separator}.

\subsubsection{Transcriber}

The Transcriber agent provides comprehensive audio-to-text transcription capabilities using advanced speech recognition models. It processes multiple audio files within a directory structure and generates accurate transcription outputs with real-time progress monitoring and robust error handling. The detailed algorithm structure is presented in Listing~\ref{lst:transcriber}.

\subsubsection{VoiceGenerator}

The VoiceGenerator agent provides advanced text-to-speech synthesis capabilities for generating high-quality voice audio from scene content. It utilizes the CosyVoice2 model for zero-shot voice cloning and produces synchronized audio with precise timestamp tracking for video editing workflows. The detailed algorithm structure is presented in Listing~\ref{lst:voicegenerator}.

\subsubsection{TTSInfer}

The TTSInfer agent provides sophisticated text-to-speech synthesis for rewritten video content, utilizing sliced audio clips as voice references. It implements a three-stage Fish-Speech pipeline combining VQGAN encoding, text-to-semantic conversion, and audio generation to produce high-quality voice clones matching original speaker characteristics. The detailed algorithm structure is presented in Listing~\ref{lst:ttsinfer}.

\subsubsection{TTSReplace}

The TTSReplace agent provides comprehensive video-audio synchronization for replacing original video audio with derivative synthesized speech segments. It implements precise temporal alignment, video speed adjustment, and seamless audio-video merging to produce coherent rewritten video content while maintaining visual continuity. The detailed algorithm structure is presented in Listing~\ref{lst:ttsreplace}.

\subsubsection{TTSSlicer}

The TTSSlicer agent provides intelligent audio segmentation for preparing audio content for text-to-speech processing and transcription workflows. It implements advanced silence detection algorithms with RMS-based energy analysis to create optimal audio chunks while maintaining temporal precision and content coherence. The detailed algorithm structure is presented in Listing~\ref{lst:ttsslicer}.

\subsubsection{SVCAnalyzer}

The SVCAnalyzer agent provides comprehensive MIDI file analysis for music cover creation workflows, extracting detailed musical information including note sequences, timing data, and tempo variations. It performs intelligent lyrics-to-music alignment by matching character counts with actual note sequences while handling rest periods and tempo changes. The detailed algorithm structure is presented in Listing~\ref{lst:svcanalyzer}.

\subsubsection{SVCConversion}

The SVCConversion agent provides intelligent audio segmentation and timestamp generation for music cover workflows, converting adapted lyrics into precisely timed JSON format suitable for video generation. It handles complex temporal alignment by parsing musical structure markers and generating seamless transition points for multimedia editing. The detailed algorithm structure is presented in Listing~\ref{lst:svcconversion}.

\subsubsection{SVCCoverist}

The SVCCoverist agent provides advanced voice timbre cloning and conversion capabilities for singing voice synthesis in music cover production. It utilizes the Seed-VC (Voice Conversion) framework to perform source-to-target vocal transformation while maintaining musical characteristics and timing precision for professional-quality audio output. The detailed algorithm structure is presented in Listing~\ref{lst:svccoverist}.

\subsubsection{SVCSingle}

The SVCSingle agent provides comprehensive singing voice synthesis for music cover production, converting adapted lyrics into high-quality vocal audio with precise temporal alignment. It utilizes the DiffSinger framework for neural singing voice generation and implements sophisticated audio processing techniques to ensure accurate timing and seamless segment concatenation. The detailed algorithm structure is presented in Listing~\ref{lst:svcsingle}.

\subsubsection{StandUpConversion}

The StandUpConversion agent provides precise temporal mapping for stand-up comedy audio segments, converting individual audio files into structured timestamp format suitable for video generation workflows. It analyzes audio duration and creates synchronized timing data for seamless multimedia editing and video synchronization. The detailed algorithm structure is presented in Listing~\ref{lst:standupconversion}.

\subsubsection{CrossTalkConversion}

The CrossTalkConversion agent provides precise temporal mapping for crosstalk audio segments, converting dual-performer dialogue into structured timestamp format suitable for video generation workflows. It analyzes audio duration and creates synchronized timing data with performer identification for seamless multimedia editing and video synchronization. The detailed algorithm structure is presented in Listing~\ref{lst:crosstalkconversion}.

\subsection{Knowledge Analysis Agents}\label{sec:knowledge}

Knowledge analysis agents process and synthesize multimedia content to generate coherent narratives and creative outputs. They combine video content extraction, narrative summarization, and rhythm-aware storyboard generation to produce structured multimedia knowledge representations. The pipeline integrates semantic understanding, temporal synchronization, and creative synthesis to transform raw content into materials ready for video production workflows.

\subsubsection{RhythmContentGenerator with Global-Aware Mechanism}

The RhythmContentGenerator extracts video segment content and creates scene-focused narrative summaries incorporating user creative requirements. Its Global-Aware Mechanism enhances video retrieval by refining raw user input into captions-aware queries in two stages: first, building a caption bank from all video content; second, integrating it with rhythm analysis to generate fine-grained, contextually grounded storyboard subqueries. The detailed algorithm structure is presented in Listing~\ref{lst:videocontentextractionagent}.

\subsubsection{NewsContentGenerator}

The NewsContentGenerator creates news summaries via a dual-agent pipeline that processes reference video transcripts and adapts them to user requirements and presentation styles, using a presenter agent for content adaptation and a judger agent for structural formatting. The detailed algorithm structure is presented in Listing~\ref{lst:newscontentgenerator}.

\subsubsection{News Presentation Style}

This subsection defines the standardized presentation methodology for video event overview content generation, ensuring consistent narrative structure and professional delivery. The complete specification is provided in Listing~\ref{lst:video_event_presentation_style}.

\subsubsection{CommentaryContentGenerator}

The CommentaryContentGenerator creates commentary content from text source materials with formatting tailored for video presentations. It employs a dual-agent architecture to transform source texts into structured video-ready narratives while preserving user-specified creative requirements. The detailed algorithm structure is presented in Listing~\ref{lst:commentarycontentgenerator}.

\subsubsection{Commentary Presentation Style}

This subsection provides guidelines for creating engaging narrative content with proper structure, pacing, and audience engagement techniques. The complete writing methodology is presented in Listing~\ref{lst:content_writing_methodology}.

\subsubsection{TTSWriter}

The TTSWriter agent rewrites transcripts for derivative video content based on sliced audio. It employs a dual-LLM approach: Claude for creative content generation and DeepSeek for precise text extraction, enabling coherent transformation while preserving original speech patterns and timing constraints. The detailed algorithm structure is presented in Listing~\ref{lst:ttswriter}.

\subsubsection{SVCAdapter}

The SVCAdapter agent adapts lyrics for music cover creation, preserving original melody structure while enabling creative content transformation. It employs multi-stage dual-LLM processing for high-quality lyrical recreation with precise character count alignment. The detailed algorithm structure is presented in Listing~\ref{lst:svcadapter}.

\subsubsection{StandUpAdapter}

The StandUpAdapter agent transforms reference content into structured stand-up comedy scripts. It employs Claude LLM for creative content generation while preserving professional comedy formatting conventions including tone markers and atmosphere cues. The detailed algorithm structure is presented in Listing~\ref{lst:standupadapter}.

\subsubsection{StandUpSynth}

The StandUpSynth agent synthesizes audio for stand-up comedy performances, combining text-to-speech generation with audience reaction integration. It employs CosyVoice2 for speech synthesis and DeepSeek LLM for script parsing to produce realistic comedy performances with appropriate vocal delivery and audience responses. The detailed algorithm structure is presented in Listing~\ref{lst:standupsynth}.

\subsubsection{CrossTalkAdapter}

The CrossTalkAdapter agent transforms reference content into traditional Chinese crosstalk (xiangsheng) dialogue format. It employs Claude LLM for cultural localization and dialogue structure optimization, preserving authentic crosstalk characteristics including role-specific delivery patterns and traditional interactive elements. The detailed algorithm structure is presented in Listing~\ref{lst:crosstalkladapter}.

\subsubsection{CrossTalkSynth}

The CrossTalkSynth agent synthesizes audio for crosstalk performances, combining dual-performer voice generation with intelligent dialogue analysis. It employs CosyVoice2 for speech synthesis and DeepSeek LLM for script parsing to produce authentic crosstalk performances with role-specific vocal characteristics. The detailed algorithm structure is presented in Listing~\ref{lst:crosstalksynth}.

\subsubsection{VideoConversion}

The VideoConversion agent transforms timestamped audio content into visual scene descriptions for video generation. It processes JSON timestamp data and converts narrative content into actionable visual scenes through content analysis. The detailed algorithm structure is presented in Listing~\ref{lst:videoconversion}.

\subsubsection{VideoQA}

The VideoQA agent analyzes video content through automated transcription and interactive question-answering. It processes multiple videos simultaneously, generates combined transcripts, and enables users to query video content via natural language. The detailed algorithm structure is presented in Listing~\ref{lst:videoqa}.

\subsubsection{VideoSummarizationGenerator}

The VideoSummarizationGenerator creates video summaries by processing video files through automatic speech recognition and applying user-specified presentation styles. It combines transcription with content adaptation to generate structured summaries. The detailed algorithm structure is presented in Listing~\ref{lst:videosummarizationgenerator}.

\subsection{Prompt of Intent Analysis}\label{sec:intent}
To reduce memory burden on LLMs and achieve precise intent-tool alignment, we conduct intent analysis on user requirements. The detailed prompt is presented in Listing~\ref{lst:intentanalysis1}. During the self-reflection phase, \model\ adaptively refines its intent analysis based on feedback; the corresponding refinement prompt is shown in Listing~\ref{lst:intentanalysis2}.

\subsection{Graph Construction Rules}\label{sec:graphconstructrule}
We provide a prompt for agent graph construction that guides collaboration among tool-use agents while allowing flexible workflow design beyond predefined pipelines, as presented in Listing~\ref{lst:agentgraphprompt}.

\subsection{Graph Evaluation}\label{sec:twostep}
During textual-gradient graph optimization, we employ a two-step evaluation strategy to maximize accuracy. The initial and secondary evaluation prompts are provided in Listing~\ref{lst:initialevaluation} and Listing~\ref{lst:secondaryevaluation}, respectively.

\subsection{Cases of Multi-modal Agent Workflow}\label{sec:casemmagentworkflow}
To demonstrate \model's sensitivity to user requirements and its generalization capability, we provide several constructed agent workflow cases. \model\ can be extensively customized to tailor workflows to user needs. For example, Listing~\ref{lst:reqaudiopreprocessing} shows a loudness normalization requirement, with the full agent graph and parameter routing provided in Listing~\ref{lst:audiopreprocessing}. Users can extend this with additional tasks such as vocal separation or background music mixing. As another illustrative case, Listing~\ref{lst:reqstorytelling} presents a storytelling video requirement, with the corresponding complete workflow shown in Listing~\ref{lst:casestorytelling}.

\subsection{Video Editing Evaluation}\label{sec:mllmbench}

In our video editing evaluation pipeline, MLLMs and VLMs serve as the core reasoning engines for clip-to-caption matching and automated content analysis, with MiniCPM-V-2.6 as the backbone.

\textbf{Clip-to-Caption Matching}. Given video clips at three-second intervals and textual captions, the VLM identifies the best-matching clip for each caption. It receives a text prompt with visual frames (1 fps) embedded as base64 in the multimodal message, and analyzes spatial features, object composition, scene layout, and temporal coherence to output a structured JSON with selections and reasoning — achieving fine-grained semantic alignment beyond keyword-based retrieval.

\textbf{Frame Integration Strategy}. Each three-second clip contributes three frames, appended sequentially after the text prompt for joint reasoning over both modalities. For longer videos, frames from all clips are packed into one multimodal request to preserve global context. The full protocol is detailed in Listing~\ref{lst:video_clip_matching_prompt}.

\textbf{Automated Content Analysis}. Beyond clip matching, the VLM assesses retrieved segments along three dimensions: semantic relevance to the user requirement, visual diversity across clips, and temporal coherence. This pipeline enables scalable, reproducible evaluation across the 2,000-sample VideoEdit benchmark, with results validated against human judgment (see Section~\ref{sec:consistency_study}).

\subsection{Artifact Licenses and Intended Use.}
Our newly constructed VideoEdit benchmark and the released VideoAgent code are intended for non-commercial research on multimodal video understanding, video editing, and agentic workflow orchestration. The artifacts should not be used for impersonation, deceptive editing, misinformation generation, or other rights-infringing or harmful content creation.

\begin{table*}[t]
\centering
\scriptsize
\setlength{\tabcolsep}{3pt}
\renewcommand{\arraystretch}{0.9}
\caption{The detailed information of tooluse agents.}
\label{tab:agentinfo}
\begin{tabularx}{\textwidth}{
    >{\raggedright\arraybackslash}p{2.6cm}
    >{\raggedright\arraybackslash}p{2.7cm}
    >{\raggedright\arraybackslash}X
}
\toprule
Agent Name & Type & Description \\
\midrule

    AudioExtractor & Audio Preprocessing & To extract audio from a single video or all videos in a directory \\
       \midrule
    LoudnessNormalizer & Audio Preprocessing & Audio loudness normalization tool \\
       \midrule
    Merge & Audio Preprocessing & To merge video and audio tracks \\
       \midrule
    Mixer & Audio Preprocessing & To mix audio with the BGM \\
       \midrule
    Resampler & Audio Preprocessing & Audio Resampling tool \\
       \midrule
    Separator & Audio Preprocessing & Audio source separation tool \\
       \midrule
    VoiceGenerator & Speech Generation & To generate speech based on scene content.\\
        \midrule
    CrossTalkAdapter & Cross-lingual Adaptations & Adapt a reference script into segmented cross talk. \\
       \midrule
    CrossTalkSynth & Cross-lingual Adaptations & Segment-by-segment cross talk audio synthesis with final merge. \\
       \midrule
    StandUpAdapter & Cross-lingual Adaptations & Adapt a reference script into segmented stand-up comedy. \\
       \midrule
    StandUpSynth & Cross-lingual Adaptations & Segment-by-segment stand-up comedy audio synthesis with final merge. \\
       \midrule
    SVCAdapter & Song Remixess & Adapt the original lyrics. \\
       \midrule
    SVCAnalyst & Song Remixess & Analyze the original song's MIDI file to extract information such as notes, note durations, etc. \\
       \midrule
    SVCAnalyst & Song Remixess & Analyze the original song's MIDI file to extract information such as notes, note durations, etc. \\
       \midrule
    SVCCoverist & Song Remixess & Source-to-target voice timbre cloning for singing voice synthesis.\\
       \midrule
    SVCSingle & Song Remixess & Split the adapted lyrics into segments, then synthesize each segment with the default vocal singing voice.\\
       \midrule
    TTSInfer & Meme Video & Take the sliced audio clips as target voice references, then conduct Text-To-Speech synthesis using their rewritten text segments.\\
       \midrule
    TTSReplace & Meme Video & Replace audio of the original video with derivative audio segments from the sliced clips.\\
       \midrule
    TTSSlicer & Meme Video & Slice the audio into segments. The audio segments often need to be transcribed afterward.\\
       \midrule
    TTSWriter & Meme Video & Rewrite the transcript of the sliced audio segments based on user requirements.\\
       \midrule
    CommentaryContent Generator & Storytelling & To generates commentary content based on user ideas and text source materials with specialized formatting for video presentations.\\
       \midrule
    NewsContentGenerator & Video Overview & To generate news summary based on user ideas and reference materials. \\
       \midrule
    RhythmDetector & Beat-synced Edits & To create cut points for video editing based on music rhythms. \\
       \midrule
    RhythmDetector & Beat-synced Edits & To create cut points for video editing based on music rhythms. \\
       \midrule
    RhythmContent Generator & Beat-synced Edits & To extract video segment content, create scene-focused narrative summaries, and generate rhythm-aware storyboards.\\
    \midrule
    VideoConversion & Video Edits & converts audio content with JSON timestamps into visual scene descriptions for video generation.\\
    \midrule
    VideoEditor & Video Edits & Ultimately merging the clips and adding audio.\\
    \midrule
    VideoPreloader & Video Edits & To initialize environment and preprocess video files.\\
    \midrule
    VideoSearcher & Video Edits & To retrieve matching video clips based on timestamp file.\\
    \midrule
    VideoQA & Video Interaction  & Transcribes all videos in a directory and provides interactive Q\&A session.\\
    \midrule
    VideoSummarization Generator & Video Interaction  & Agent that generates summarization content based on user ideas and reference materials.\\
    \midrule
    FaceSwapping & Video Processing & Replace the faces appearing in the video with those of the target person.\\
    \midrule
    LipSynchronization & Video Processing & Synchronize the speaker's lip movements in the video with the audio track.\\
\bottomrule
\end{tabularx}
\end{table*}

\lstset{
    basicstyle=\ttfamily\scriptsize,
    breaklines=true,
    columns=fullflexible,
    keepspaces=true,
    frame=single,
    framesep=5pt,
    xleftmargin=0pt,
    breakindent=0pt,
    xrightmargin=5pt,
    showstringspaces=false,
    aboveskip=2pt,
    belowskip=2pt
}

\begin{figure*}[t]
\centering
\begin{minipage}{\textwidth}
\begin{lstlisting}[
    caption={VideoAgent Showcase 1 with Prompts},
    label=lst:videoagent-showcase1
]
Categories:
1. Beat-synced Edits
   Description:
     - User provides background music and film footage.
     - Detects beat/tempo changes and aligns high-energy shots to strong beats.
   Prompt:
    Begin with Gwen, who has blond hair, sitting at a dining table in front of a window, then transition to her playing drums with pop textures and musical notes in the background. Include action sequences featuring Miguel O'Hara in a dark blue suit with red accents, sharp red claws, and black-and-red eye lenses; Spider-Gwen in a white-and-pink suit with a hood and ballet shoes; Miles Morales with curly hair and a red spider logo on his chest; and the Spot in a black suit covered in white spots using portal powers. Focus on the chase scene against a blue sky with trains, and emphasize high-quality motion throughout--web-swinging, combat, and vibrant special effects.


2. Video Overviews
   Description:
     - User provides a news/event video.
     - Transcribe speech, summarize, align summary sentences to visuals, generate voiceover.
   Prompt:
     - Short tech news, colloquial expression within 250 words, check the accuracy of key terms, e.g. the GPT model name should be 4o instead of 4.0

3. Storytelling
   Description:
     - User provides story/novel text and footage.
     - Script in chosen narration style, match lines to visuals, generate voiceover.
   Prompt:
     - A verbal interpretation copy of no less than 1,000 words

4. Meme Video Edits
   Description:
     - User supplies a source video and custom script.
     - Extract/transcribe audio, generate new speech, sync precisely to frames, output dubbed edit.
   Prompt:
     - Create a humorous narrative about two PhD students seeking advice from Master Ma. For the two PhD students, one of them is known for high citation counts and the other for numerous publications. Transform martial arts terms into AI research terminology while keeping phrase lengths similar (length difference should be less than two Chinese characters). The story highlights their academic rivalry and ends with Master Ma advising against "internal competition". Keep signature phrases like "wasn't cautious enough" and "achieving great results with minimal effort" while avoiding mentions of real institutions. The word combinations should be logical and appropriate for an academic context.

5. Song Remixes
   Description:
     - User provides MIDI, lyrics, background music, and a target voice sample.
     - Generate cover, clone voice, align timing, integrate into video pipeline.
   Prompt:
     - The song is performed by Patrick Star, focusing on the theme of "the struggles of manuscript submission and dealing with overly critical reviewers", following the original lyrics' sentence structure while replacing specific content. It incorporates elements of reviewer nitpicking (e.g., questioning innovation, demanding redundant experiments) and expresses frustration with lines like "If only I could swap reviewers, this academic fate is too cruel" to highlight the emotional toll of peer review.

6. Cross-lingual Adaptations
   Description:
     - User provides source audio and target voice samples.
     - Adapt script to cultural format (e.g., crosstalk/talk show), synthesize voices, apply effects, integrate into pipeline.
   Prompt:
     - Adapting to the cultural context of talk shows and localizing humor
\end{lstlisting}
\end{minipage}
\end{figure*}

\begin{figure*}[t]
\centering
\begin{minipage}{\textwidth}
\begin{lstlisting}[
    caption={VideoAgent Showcase 2 with Prompts},
    label=lst:videoagent-showcase2
]
Categories:
1. Beat-synced Edits
   Description:
     - User provides background music and film footage.
     - Detects beat/tempo changes and aligns high-energy shots to strong beats.
   Prompt:
     - Conflicts between Nezha, Dragon Prince Ao Bing (blue-robed and blue hair) and Shen Gongbao (black-robed).

2. Video Overviews
   Description:
     - User provides a news/event video.
     - Transcribe speech, summarize, align summary sentences to visuals, generate voiceover.
   Prompt:
     - Short movie podcast, colloquial expression within 300 words; identify which actor or host is speaking; do not mention movie-ticket availability.

3. Storytelling
   Description:
     - User provides story/novel text and footage.
     - Script in chosen narration style, match lines to visuals, generate voiceover.
   Prompt:
     - Write a fluent commentary script of 1,500 words.

4.1 Meme Video Edits (Fan Zhiyi)
   Description:
     - User supplies a source video and custom script.
     - Extract/transcribe audio, generate new speech, sync precisely to frames, output dubbed edit.
   Prompt:
     - Emphasize the positive impact of IShowSpeed's China tour, which is not only an online carnival but also a milestone event in cultural communication in the digital age.

4.2 Meme Video Edits (Xiao Ming Live)
   Description:
     - User supplies a source video and custom script.
     - Extract/transcribe audio, generate new speech, sync precisely to frames, output dubbed edit.
   Prompt:
     - Based on the following scenario, create an angry rebuttal from Zhuge Liang (me):
       - Speaker: Zhuge Liang (me)
       - Start with "Why don't you look at your own problems for the failure," followed by "...look at your own problems" pattern sentences that all reference anime events
       - Anime examples must mention specific characters
       - Only the last "...look at your own problems" should return to the failure scenario
       - Use colloquial language and diverse anime references

5. Cross-lingual Adaptations
   Description:
     - User provides source audio and target voice samples.
     - Adapt script to cultural format (e.g., crosstalk/talk show), synthesize voices, apply effects, integrate into pipeline.
   Prompt:
     - To generate a crosstalk-style piece, the story must conform to objective reality and be about forty to fifty sentences.
\end{lstlisting}
\end{minipage}
\end{figure*}

\begin{figure*}[t]
\centering
\begin{minipage}{\textwidth}
\begin{lstlisting}[
    caption={VideoAgent Showcase 3 with Prompts},
    label=lst:videoagent-showcase3
]
Categories:
1.1 Beat-synced Edits (Titanic)
   Description:
     - User provides background music and film footage.
     - Detects beat/tempo changes and aligns high-energy shots to strong beats.
   Prompt:
     - A romantic and sweet love story about Jack and Rose meeting on the Titanic. It cannot include the part where the ship is in distress, nor the night scene. In the first section, Rose, wearing a purple hat and a white shirt, walks out of a white car with a purple umbrella, looking thoughtfully.

1.2 Beat-synced Edits (Interstellar)
   Description:
     - User provides background music and film footage.
     - Detects beat/tempo changes and aligns high-energy shots to strong beats.
   Prompt:
     - Celebrate humanity's courage in space exploration. Include scenes featuring spaceships, wormholes, black holes, space station docking maneuvers, ocean planets, and glacial worlds. Show astronauts in their distinctive white spacesuits as they venture into the unknown, highlighting humanity's drive to explore the cosmos.

1.3 Beat-synced Edits (Interstellar)
   Description:
     - User provides background music and film footage.
     - Detects beat/tempo changes and aligns high-energy shots to strong beats.
   Prompt:
     - Love can transcend time and space.

1.4 Beat-synced Edits (Ne Zha)
   Description:
     - User provides background music and film footage.
     - Detects beat/tempo changes and aligns high-energy shots to strong beats.
   Prompt:
     - Capture more scenes of conflicts and battles between Nezha and Shen Gongbao (black-robed) and Dragon Prince Ao Bing (blue-robed).

2.1 Meme Video Edits (Xiao Ming Live)
   Description:
     - User supplies a source video and custom script.
     - Extract/transcribe audio, generate new speech, sync precisely to frames, output dubbed edit.
   Prompt:
     - Background: Mixue Ice Cream is a national chain brand focusing on ice cream and tea beverages. On March 15 (Consumer Rights Day), they were reported to be using overnight lemons. However, compared to other exposures, using overnight lemons is not considered a particularly serious violation and is somewhat understandable.
     - Speaker: Snow King (Mixue's representative)
     - Purpose: Emphasize that the "overnight lemon" situation is not too serious, highlighting Mixue's good reputation
     - Must preserve the phrases "Look in my eyes tell me why" and "tell me"
     - Must end with the word "say it"
     - If the original text contains awkward phrasing, such as redundant words or confused semantics, do not imitate that style or sentence structure
     - Ensure natural and fluent sentences

2.2 Meme Video Edits (Xiao Ming Live)
   Description:
     - User supplies a source video and custom script.
     - Extract/transcribe audio, generate new speech, sync precisely to frames, output dubbed edit.
   Prompt:
     - Based on the following scenario, create an angry rebuttal from Zhuge Liang (me):
       - Speaker: Zhuge Liang (me)
       - Zhuge Liang (me) is challenged about why a certain Three Kingdoms character has a higher rating than him and launches a fierce rebuttal
       - Later rating comparisons should show stark differences (can be exaggerated)
       - Use colloquial language, align with historical facts, and only replace specific content
\end{lstlisting}
\end{minipage}
\end{figure*}

\begin{figure*}[t]
\centering
\begin{minipage}{\textwidth}
\begin{lstlisting}[
    caption={Audio Editing Instruction Prompt},
    label=lst:audio_editing_instruction_prompt
]
You are an Autonomous Agent System Designer. I will provide Registered Agents (Name, Description and Parameters information):

Your task is to:
1. Generate Feasible User Requirement:
   - Only consider generating feasible User Requirement for **audio-related** aspects (Finally generate audio instead of video) based on metadata of Registered Agents.
   - Avoid mentioning specific agent names in User Requirement, but clearly describe the needs.
   - Avoid mentioning the concrete steps and details of implementation.
   - Reflect real-world needs, and be phrased conversationally.
2. Design Executable Agent Graph:
   - Format: List
   - Agent Graph shall contain metadata for each Agent Node including:
     * name: (string)
     * inputs: (list of input parameter objects with):
       - parameter: input parameter name
       - description: brief parameter description
     * outputs: (list of output parameter objects with):
       - parameter: output parameter name
       - description: brief parameter description
       - links: (list of dictionaries) where each dictionary specifies:
         - key of dictionaries: target agent name
         - value of dictionaries: target agent's input parameter name that this output connects to
   - Agent Graph case:
       [
          {
           "node": Agent Name,
           "inputs": [
                {
                "name": input parameter name1,
                "description": ...
                },
                {
                "name": input parameter name2,
                "description": ...
                },
              ]
           "outputs": [
                {
                "name": output parameter name1,
                "description": ...,
                "links": [
                    {"next_agent1": next_agent1's input parameter name"},
                    {"next_agent2": next_agent2's input parameter name"}
                    ]
                },
                {
                "name": output parameter name2,
                "description": ...,
                "links": []
                },
                ...
              ]
           },
           ...
       ]
3. Generate Agent Chain:
   - Format: List
   - Based on the description of the Agent and the sequential information contained in the designed Agent Graph, generate the Agent Chain
   - Agent Chain case:
        ["agent1", "agent2", ...]
4. Generate User Input Graph
   - Format: List
   - Parameter nodes with no in-degree (no incoming edges) are uniformly considered to require user input.
   - Parameter nodes with no in-degree may have different names but share the same user input, meaning a single user input parameter can point to multiple such nodes.
   - Parameter nodes with no in-degree that are linked to user input should be represented in the format **AgentName.input_parameter**
   - Generate the User Input Graph based on the Agent descriptions and parameter passing information in the designed Agent Graph.
   - User Input Graph case:
       [
         {
          "node": User input parameter name,
          "description": brief description of the parameter
\end{lstlisting}
\end{minipage}
\end{figure*}
\begin{figure*}[t]
\centering
\begin{minipage}{\textwidth}
\noindent\textbf{Listing~\ref{lst:audio_editing_instruction_prompt} continued: Audio Editing Instruction Prompt.}
\begin{lstlisting}
          "links": [
                    {"agent1": agent1.input_parameter1},
                    {"agent2": agent2.input_parameter2}
               ]
         },
         ...
       ]
5. Output Reasoning:
   - Provide concise reasoning (<200 words) explaining the entire workflow logic

In addition to the above formatting requirements, please also note the following:
1. For each element of **outputs** in each Agent Node:
   - Ensure that the **links** in the **outputs** point to an input parameter that actually exists in the next Agent Node.
   - Also, make sure the description of the output parameter matches the description of the input parameter in the next Agent Node.
2. Final JSON Output Format Specification:
{
    "User Requirement": ...,
    "Agent Graph": ...,
    "Agent Chain": ...,
    "User Input Graph": ...,
    "Reasoning": ...
}
Strictly follow JSON output format!

Metadata of registered agents:
{registry_agents}

Real-life Requirement Examples:
{real-life_examples}

\end{lstlisting}
\end{minipage}
\end{figure*}

\begin{figure*}[t]
\centering
\begin{minipage}{\textwidth}
\begin{lstlisting}[
    caption={Video Editing Instruction Prompt},
    label=lst:video_editing_instruction_prompt
]
You are an Autonomous Agent System Designer. I will provide Registered Agents (Name, Description and Parameters information):

Your task is to:
1. Generate Feasible User Requirement:
   - Only consider generating feasible User Requirement for **video-related** aspects based on metadata of Registered Agents.
   - Avoid mentioning specific agent names in User Requirement, but clearly describe the needs.
   - Avoid mentioning the concrete steps and details of implementation.
   - Reflect real-world needs, and be phrased conversationally.
2. Design Executable Agent Graph:
   - Format: List
   - Agent Graph shall contain metadata for each Agent Node including:
     * name: (string)
     * inputs: (list of input parameter objects with):
       - parameter: input parameter name
       - description: brief parameter description
     * outputs: (list of output parameter objects with):
       - parameter: output parameter name
       - description: brief parameter description
       - links: (list of dictionaries) where each dictionary specifies:
         - key of dictionaries: target agent name
         - value of dictionaries: target agent's input parameter name that this output connects to
   - Agent Graph case:
       [
          {
           "node": Agent Name,
           "inputs": [
                {
                "name": input parameter name1,
                "description": ...
                },
                {
                "name": input parameter name2,
                "description": ...
                },
              ]
           "outputs": [
                {
                "name": output parameter name1,
                "description": ...,
                "links": [
                    {"next_agent1": next_agent1's input parameter name"},
                    {"next_agent2": next_agent2's input parameter name"}
                    ]
                },
                {
                "name": output parameter name2,
                "description": ...,
                "links": []
                },
                ...
              ]
           },
           ...
       ]
3. Generate Agent Chain:
   - Format: List
   - Based on the description of the Agent and the sequential information contained in the designed Agent Graph, generate the Agent Chain
   - Agent Chain case:
        ["agent1", "agent2", ...]
4. Generate User Input Graph
   - Format: List
   - Parameter nodes with no in-degree (no incoming edges) are uniformly considered to require user input.
   - Parameter nodes with no in-degree may have different names but share the same user input, meaning a single user input parameter can point to multiple such nodes.
   - Parameter nodes with no in-degree that are linked to user input should be represented in the format **AgentName.input_parameter**
   - Generate the User Input Graph based on the Agent descriptions and parameter passing information in the designed Agent Graph.
   - User Input Graph case:
       [
         {
          "node": User input parameter name,
          "description": brief description of the parameter
\end{lstlisting}
\end{minipage}
\end{figure*}
\begin{figure*}[t]
\centering
\begin{minipage}{\textwidth}
\noindent\textbf{Listing~\ref{lst:video_editing_instruction_prompt} continued: Video Editing Instruction Prompt.}
\begin{lstlisting}
          "links": [
                    {"agent1": agent1.input_parameter1},
                    {"agent2": agent2.input_parameter2}
               ]
         },
         ...
       ]
5. Output Reasoning:
   - Provide concise reasoning (<200 words) explaining the entire workflow logic

In addition to the above formatting requirements, please also note the following:
1. For each element of **outputs** in each Agent Node:
   - Ensure that the **links** in the **outputs** point to an input parameter that actually exists in the next Agent Node.
   - Also, make sure the description of the output parameter matches the description of the input parameter in the next Agent Node.
2. Final JSON Output Format Specification:
{
    "User Requirement": ...,
    "Agent Graph": ...,
    "Agent Chain": ...,
    "User Input Graph": ...,
    "Reasoning": ...
}
Strictly follow JSON output format!

Metadata of registered agents:
{registry_agents}

Real-life Requirement Examples:
{real-life_examples}
\end{lstlisting}
\end{minipage}
\end{figure*}

\begin{figure*}[t]
\centering
\begin{minipage}{\textwidth}
\begin{lstlisting}[
    caption={Video Caption to High-Level Summary Transformation Prompt},
    label=lst:video_caption_transform_prompt
]
prompt = f"""
Transform the following detailed video caption into a concise, high-level summary that captures the main themes, setting, and overall atmosphere in no more than 50 words, no point form, use pure sentence.

Focus on:
- Setting identification: Where does the action take place?
- Key themes: What are the main activities or purposes being shown?
- Atmosphere and tone: What mood or feeling does the scene convey?
- Overall narrative: What story or message emerges?

Instructions:
1. Condense specific details into broader concepts
2. Identify recurring elements that suggest themes
3. Synthesize individual actions into a cohesive narrative about the environment
4. Use descriptive language that captures essence rather than listing events
5. Maintain the core message while abstracting from granular details
6. Preserve emotional tone and workplace/environmental culture
7. Keep response under 50 words

Input format: whole_caption"[detailed description]"
Output format: high_level_caption"[concise thematic summary]"

Rules:
- Extract thematic elements from specific actions
- Focus on overarching narrative rather than individual events
- Capture the purpose
- Ensure output maintains fidelity to original content intent

Transform the provided detailed caption following these guidelines."""
\end{lstlisting}
\end{minipage}
\end{figure*}

\clearpage

\begin{figure*}[t]
\centering
\begin{minipage}{\textwidth}
\begin{lstlisting}[
    caption={VideoPreloader Structure},
    label=lst:videopreloader
]
Input: video_directory
Output: preprocessing_status

Algorithm:
1. Configure multiprocessing (spawn), logging, project paths; add modules to system path.
2. Create directory structure: audio\_analysis, scene\_output, videosource-workdir, writing\_data, video\_output; set VideoRAG working directory.
3. Validate video source directory; scan for MP4 files; report discovery statistics.
4. Dynamically import VideoRAG and QueryParam for content indexing; handle import errors with fallbacks.
5. Execute video insertion via VideoRAG: extract metadata/features, generate content embeddings, build searchable index database.
6. Report processing statistics and return execution status for downstream pipeline coordination.

Error Handling: directory creation failures, VideoRAG import errors, file access/format validation, graceful cleanup on interruption.
\end{lstlisting}
\end{minipage}
\end{figure*}

\begin{figure*}[t]
\centering
\begin{minipage}{\textwidth}
\begin{lstlisting}[
    caption={VideoSearcher Structure},
    label=lst:videosearcher
]
Input: video_scene_path
Output: search_status

Algorithm:
1. Initialize dataset, scene\_output, and working directory paths; create missing directories; add tools to system path.
2. Dynamically import VideoRAG and QueryParam; handle import failures with error logging.
3. Load scene JSON (UTF-8), extract segment\_scene as query string, validate content availability.
4. Configure QueryParam (videoragcontent mode); toggle reference inclusion via wo\_reference flag; set multiprocessing spawn.
5. Execute semantic query via VideoRAG: embed scene descriptions, similarity-match against indexed videos, retrieve relevant segments with timestamps.
6. Validate query completion, log search statistics, return structured search status with error reporting.

Error Handling: JSON parse/not-found errors, VideoRAG import/init failures, empty scene content, query exceptions with logging.
\end{lstlisting}
\end{minipage}
\end{figure*}

\begin{figure*}[t]
\centering
\begin{minipage}{\textwidth}
\begin{lstlisting}[
    caption={VideoEditor Structure},
    label=lst:videoeditor
]
Input: video_directory, audio_path, timestamp_path
Output: final_video_path

Algorithm:
1. Load MiniCPM-V-2\_6-int4 model; parse visual\_retrieved\_segments.json and kv\_store\_video\_segments.json for segment and timing metadata.
2. Parse rhythm\_points.json for beat timestamps; create time periods from consecutive beat intervals; load video\_scene.json for storyboards and align with narrative segments.
3. For each video segment, extract frames at 1fps (224x224 RGB PIL Images), starting from segment start time plus subsequent whole seconds.
4. Multimodal scene matching: construct VLM message with prompt text as first content element, then append extracted frames sequentially. The prompt asks the model to find the optimal starting frame for the best consecutive sequence matching the scene description, returning only the starting frame number.
5. Parse VLM response (regex-based number extraction); compute clip timing (start\_time + frame\_offset); validate boundaries; extract clips matching beat intervals.
6. Audio integration: load and resize BGM; composite with original audio (keep\_original\_audio=True) or replace entirely (False); adjust volume and sync timeline.
7. Concatenate clips via compose; export with fps=24, codec=libx264, audio\_codec=aac, preset=medium, threads=4; cleanup temp resources.

Error Handling: frame extraction failures, VLM parsing errors (regex fallback), file access errors, audio sync fallbacks, memory cleanup.
\end{lstlisting}
\end{minipage}
\end{figure*}

\begin{figure*}[t]
\centering
\begin{minipage}{\textwidth}
\begin{lstlisting}[
    caption={FaceSwapping Agent Structure},
    label=lst:faceswapping
]
Input: source_video_path, target_face_image
Output: swapped_video_path

Algorithm:
1. Initialization:
   a. Set up file paths and working directories
   b. Initialize access to Viggle AI face swapping tool

2. Source video processing:
   a. Load source video and extract frames
   b. Detect and track the specified face across frames
   c. Prepare frames containing the target face for swapping

3. Face swapping operation:
   For each frame with detected target face:
     a. Send frame and target_face_image to Viggle AI tool
     b. Receive the frame with the specified face replaced by the target face
     c. Replace the original frame with the swapped output

4. Video reconstruction:
   a. Reassemble processed frames preserving original video properties
   b. Synchronize original audio track if available
   c. Output the final face-swapped video file

5. Error handling and cleanup:
   a. Handle failures in calling Viggle AI or frame processing gracefully
   b. Log issues and fall back to original frames if needed
   c. Release resources and temporary files

Notes:
- Input video and target face image are required inputs
- Output is a video with the specified face replaced by the target
- Reliant on Viggle AI for core face swapping capabilities
- Designed for seamless integration within video pipelines
\end{lstlisting}
\end{minipage}
\end{figure*}

\begin{figure*}[t]
\centering
\begin{minipage}{\textwidth}
\begin{lstlisting}[
    caption={LipSynchronization Agent Structure},
    label=lst:lipsynchronization
]
Input: source_video_path, target_audio_path
Output: synced_video_path

Algorithm:
1. Initialization:
   a. Configure input/output file paths and working directories
   b. Initialize connection and access to Kling AI lip synchronization tool

2. Data preparation:
   a. Load source video and extract frames and original audio if present
   b. Load target audio track for synchronization

3. Lip synchronization process:
   a. Send source video frames and target audio to Kling AI tool
   b. Receive processed video frames with speaker's lip movements synchronized to target audio
   c. Replace original frames with synchronized frames

4. Video reconstruction:
   a. Combine synchronized frames preserving original video resolution and frame rate
   b. Integrate the target audio track as the output video's audio
   c. Output final lip-synced video file

5. Error handling and cleanup:
   a. Handle communication errors with Kling AI tool gracefully
   b. Provide fallback to original video if synchronization fails
   c. Log all processing steps and encountered issues
   d. Release temporary resources and close file streams

Notes:
- Supports video and separate audio inputs
- Output video shows visually consistent lip motions aligned with target audio
- Relies fully on Kling AI external service for synchronization
- Designed for smooth integration with video editing workflows
\end{lstlisting}
\end{minipage}
\end{figure*}

\begin{figure*}[t]
\centering
\begin{minipage}{\textwidth}
\begin{lstlisting}[
    caption={AudioExtractor Structure},
    label=lst:audioextractor
]
Input: video_path (file or directory)
Output: audio_paths, data_directory

Algorithm:
1. Validate video\_path (file or directory); scan for supported formats (.mp4, .avi, .mov, .mkv, .wmv, .flv, .webm, .m4v).
2. For each video, extract audio via FFmpeg: \texttt{ffmpeg -y -i input -vn -acodec pcm\_s16le -ar 44100 -ac 2 -loglevel error output.wav}. Handle subprocess errors (CalledProcessError, FileNotFoundError for FFmpeg).
3. Batch processing: track success/failure per file, generate progress reports, collect successful paths, report statistics.
4. Return audio\_paths (list for batch, string for single) and data\_dir; handle empty results gracefully.

Error Handling: FFmpeg availability detection, per-file error isolation, directory permission validation, graceful degradation for partial failures.
\end{lstlisting}
\end{minipage}
\end{figure*}

\begin{figure*}[t]
\centering
\begin{minipage}{\textwidth}
\begin{lstlisting}[
    caption={RhythmDetector Structure},
    label=lst:rhythmdetector
]
Input: audio_file_path
Output: rhythm_analysis_directory

Algorithm:
1. Load audio via librosa.load() at original sample rate; configure STFT parameters (frame\_length=2048, hop\_length=512).
2. Compute RMS energy (librosa.feature.rms), normalize (rms/max), smooth via convolution kernel.
3. Detect peaks via scipy.signal.find\_peaks (height=0.4 threshold, configurable distance); convert indices to timestamps (librosa.frames\_to\_time); apply temporal masking for intro/outro exclusion.
4. Generate rhythm points with sequential IDs and precise timing; filter masked regions; report statistics.
5. Create 3-panel visualization (waveform + markers, RMS curve + threshold, spectrogram) at 300 DPI with color-coded rhythm points (red) and masked areas (gray).
6. Compute inter-beat interval statistics (mean, median, std, min, max); generate distribution histograms.
7. Export cut\_points.json (beat timing), rhythm\_detection.png, and rhythm\_distribution.png to audio\_analysis/.

Error Handling: format validation, path verification, graceful handling of insufficient rhythm points, file I/O exceptions.
\end{lstlisting}
\end{minipage}
\end{figure*}

\begin{figure*}[t]
\centering
\begin{minipage}{\textwidth}
\begin{lstlisting}[
    caption={LoudnessNormalizer Structure},
    label=lst:loudnessnormalizer
]
Input: data_directory
Output: processing_status

Algorithm:
1. Validate and resolve input directory path via Path.resolve(); verify directory exists and is accessible.
2. Build FAP command: \texttt{fap loudness-norm <input\_dir> <output\_dir> --overwrite --recursive}.
3. Execute via subprocess.Popen with stdout/stderr PIPE (bufsize=1, line-buffered); spawn threads for real-time UTF-8 output reading.
4. Monitor progress with [FAP]-prefixed console output; handle decoding errors with 'replace'.
5. Wait for completion (process.wait), join threads, evaluate return\_code: return success for 0, raise RuntimeError otherwise.

Error Handling: path validation, FAP availability detection, thread synchronization, subprocess error capture, resource cleanup.
\end{lstlisting}
\end{minipage}
\end{figure*}

\clearpage

\begin{figure*}[t]
\centering
\begin{minipage}{\textwidth}
\begin{lstlisting}[
    caption={Merge Structure},
    label=lst:merge
]
Input: video_path, audio_path
Output: merged_video_path

Algorithm:
1. Validate video\_path and audio\_path; set output to overwrite original video.
2. Execute FFmpeg merge: \texttt{ffmpeg -i video\_path -i audio\_path -c:v libx264 -preset fast -crf 23 -c:a aac -b:a 192k -map 0:v:0 -map 1:a:0 -shortest -movflags +faststart -y output\_path}.
3. Monitor return code; validate output file creation; ensure A/V sync.
4. Handle errors (CalledProcessError, permission, cleanup) with detailed reporting.
\end{lstlisting}
\end{minipage}
\end{figure*}

\begin{figure*}[t]
\centering
\begin{minipage}{\textwidth}
\begin{lstlisting}[
    caption={Mixer Structure},
    label=lst:mixer
]
Input: bgm_path, audio_path
Output: mixed_audio_path

Algorithm:
1. Validate BGM and audio paths; determine output format from extension (default WAV); set output path as "mixed\_" + original name.
2. Load BGM and vocal via AudioSegment.from\_file (supports WAV, MP3, FLAC etc.); apply BGM volume reduction (default -1 dB).
3. If BGM is shorter than vocal, loop-extend BGM to vocal length, then trim to exact match.
4. Apply overlay: vocal\_audio.overlay(bgm) for vocal-priority mixing; preserve audio quality.
5. Export via mixed\_audio.export(output\_path, format) (WAV/MP3/FLAC/AAC/OGG); return output path.

Error Handling: format compatibility, loading errors, export fallback to WAV.
\end{lstlisting}
\end{minipage}
\end{figure*}

\begin{figure*}[t]
\centering
\begin{minipage}{\textwidth}
\begin{lstlisting}[
    caption={Resampler Structure},
    label=lst:resampler
]
Input: data_directory
Output: processing_status

Algorithm:
1. Validate and resolve input directory via Path.resolve(); verify exists and is a directory.
2. Build FAP command: \texttt{fap resample <input\_dir> <output\_dir> --overwrite} (in-place replacement).
3. Execute via subprocess.Popen (stdout/stderr PIPE, bufsize=1); spawn threads for real-time UTF-8 output reading with [FAP] prefix.
4. Wait for completion, join threads; return success for return\_code=0, raise RuntimeError otherwise.

Error Handling: path validation, FAP availability, thread synchronization, return code analysis.
\end{lstlisting}
\end{minipage}
\end{figure*}

\begin{figure*}[t]
\centering
\begin{minipage}{\textwidth}
\begin{lstlisting}[
    caption={Separator Structure},
    label=lst:separator
]
Input: data_directory
Output: processing_status

Algorithm:
1. Validate and resolve input directory; verify exists and is a directory type.
2. Build FAP command: \texttt{fap separate <input\_dir> <output\_dir> --overwrite --recursive} (ML-based source separation).
3. Execute via subprocess.Popen with threaded stdout/stderr monitoring; real-time UTF-8 output with [FAP] prefix.
4. Wait for completion, join threads; evaluate return\_code; generate separated vocal and instrumental tracks.

Error Handling: directory validation, FAP availability, thread sync, resource cleanup for large files.
\end{lstlisting}
\end{minipage}
\end{figure*}

\begin{figure*}[t]
\centering
\begin{minipage}{\textwidth}
\begin{lstlisting}[
    caption={Transcriber Structure},
    label=lst:transcriber
]
Input: data_directory
Output: processing_status

Algorithm:
1. Validate and resolve input directory; verify exists and is a directory type.
2. Build FAP command: \texttt{fap transcribe --model-type funasr --recursive <audio\_dir>}.
3. Execute via subprocess.Popen with threaded stdout/stderr monitoring; real-time UTF-8 output with [FAP] prefix.
4. Wait for completion, join threads; evaluate return\_code; generate .txt files matching audio basenames.

Error Handling: directory validation, FunASR availability, format compatibility, thread cleanup.
\end{lstlisting}
\end{minipage}
\end{figure*}

\begin{figure*}[t]
\centering
\begin{minipage}{\textwidth}
\begin{lstlisting}[
    caption={VoiceGenerator Structure},
    label=lst:voicegenerator
]
Input: video_scene_path, target_vocal_path
Output: synthesized_audio_path, timestamp_path

Algorithm:
1. Load CosyVoice2-0.5B model (load\_jit/load\_trt=False, fp16=False); load target vocal prompt at 16kHz via load\_wav.
2. Parse scene JSON (UTF-8, ensure\_ascii=False); extract content\_created via '/////' delimiter; create segment list with sequential IDs.
3. Split text into max 200-char chunks; handle Chinese and bilingual punctuation; maintain semantic coherence.
4. For each chunk, apply zero-shot synthesis: cosyvoice.inference\_zero\_shot(generator, prompt\_text, prompt\_speech\_16k, stream=False); concatenate via torch.cat(); track cumulative timestamps.
5. Export final audio via torchaudio.save(); generate cut\_points.json with sentence\_data.chunks structure; cleanup temp files.

Error Handling: model loading verification, file validation, chunk-level error isolation, resource cleanup.
\end{lstlisting}
\end{minipage}
\end{figure*}

\begin{figure*}[t]
\centering
\begin{minipage}{\textwidth}
\begin{lstlisting}[
    caption={TTSInfer Structure},
    label=lst:ttsinfer
]
Input: audio_path, speech_text_path
Output: derivative_audio_directory

Algorithm:
1. Load rewritten text (UTF-8), split into paragraphs; scan slice directory for .lab/.wav pairs; create derivative output directory.
2. For each paragraph, map .lab content to .wav reference audio; aggregate additional segments via modulo cycling if combined duration < 1s; concatenate WAVs via scipy.io.wavfile.
3. Three-stage Fish-Speech pipeline: (1) VQGAN encode reference WAV -> .npy tokens (checkpoint: firefly-gan-vq-fsq); (2) Text-to-Semantic conversion with --text and --prompt-text; (3) VQGAN decode semantic codes -> synthesized WAV.
4. Process each paragraph through all 3 stages; validate return codes per stage; collect WAVs.
5. Sort, load, and concatenate via numpy.concatenate(); export derivative/final.wav.

Error Handling: return code validation per stage, sample rate consistency, temp file cleanup.
\end{lstlisting}
\end{minipage}
\end{figure*}

\begin{figure*}[t]
\centering
\begin{minipage}{\textwidth}
\begin{lstlisting}[
    caption={TTSReplace Structure},
    label=lst:ttsreplace
]
Input: video_path
Output: final_video_path

Algorithm:
1. Validate video path; load metadata.json (clip file names, start/end times, durations); map derivative audio to video segments.
2. Per clip: extract video via FFmpeg (\texttt{-ss start -to duration -c:v libx264 -preset fast -vf scale=iw:ih}); get audio duration via FFprobe; compute speed factor (target/clip duration).
3. Adjust video speed via setpts filter (\texttt{-filter:v setpts=speed\_factor*PTS -an}); merge with derivative audio (\texttt{-c:v copy -c:a aac -map 0:v:0 -map 1:a:0 -shortest}).
4. Generate filelist.txt (UTF-8, absolute paths); concatenate all clips via FFmpeg concat demuxer (\texttt{-f concat -safe 0 -c copy}); cleanup temp files, keep only final.mp4.

Error Handling: file validation, subprocess return codes, missing audio segments, resource cleanup.
\end{lstlisting}
\end{minipage}
\end{figure*}

\clearpage

\begin{figure*}[t]
\centering
\begin{minipage}{\textwidth}
\begin{lstlisting}[
    caption={TTSSlicer Structure},
    label=lst:ttsslicer
]
Input: audio_path
Output: processing_status, segmented_audio_files, metadata

Algorithm:
1. Configure parameters: min/max segment duration (6s/8s), silence threshold (top\_db=-35), hop\_length=10ms, max\_silence\_kept=0.3s.
2. Load audio via librosa; compute RMS energy (frame-based, librosa.feature.rms) with dB-to-linear threshold conversion.
3. Detect silence regions (RMS < threshold); apply leading/middle silence segmentation rules; compute optimal cut points via argmin on RMS; generate sil\_tags array.
4. Extract chunks via \_apply\_slice: convert frame indices to sample indices, slice waveform, compute precise timestamps (start\_idx/sr, end\_idx/sr).
5. Post-process: merge short chunks or subdivide long ones via \_slice\_by\_max\_duration (ceil for uniform distribution).
6. Export via soundfile.write (sequential numbering: 0000.wav, ...); generate metadata.json (UTF-8) with start/end timestamps (3-decimal precision).

Error Handling: format validation, empty audio detection, sample rate preservation, file I/O for large files.
\end{lstlisting}
\end{minipage}
\end{figure*}

\begin{figure*}[t]
\centering
\begin{minipage}{\textwidth}
\begin{lstlisting}[
    caption={SVCAnalyzer Structure},
    label=lst:svcanalyzer
]
Input: midi_file_path, lyrics_file_path
Output: song_name, analysis_results_path

Algorithm:
1. Validate MIDI (.mid) and lyrics (.txt) files; load lyrics (UTF-8); extract song name.
2. Parse MIDI via mido.MidiFile; scan tracks for 'set\_tempo' messages; compute BPM (60000000/tempo); handle variable tempo with chronological sorting.
3. Extract notes: process note\_on (velocity>0) and note\_off; compute duration in ticks; detect rests (gaps > ticks\_per\_beat/8); group simultaneous notes (<0.01 time units).
4. Convert ticks to seconds via tempo-aware formula: (ticks * tempo) / (ticks\_per\_beat * 1e6); format note names as "{note}{octave}" (e.g., "C4", "F#3").
5. Align lyrics character-by-character to notes (excluding rests); insert "AP" markers for rest periods; generate processed lyrics.
6. Export JSON to /analysis/{song\_name}.json with fields: text (with AP markers), notes (pipe-separated), notes\_duration, input\_type="word".

Error Handling: MIDI/lyrics validation, encoding fallbacks, character count mismatch detection.
\end{lstlisting}
\end{minipage}
\end{figure*}

\begin{figure*}[t]
\centering
\begin{minipage}{\textwidth}
\begin{lstlisting}[
    caption={SVCConversion Structure},
    label=lst:svcconversion
]
Input: adapted_lyrics_string, midi_analysis_path
Output: timestamp_json_path

Algorithm:
1. Load MIDI analysis JSON (UTF-8); extract note durations; update with adapted lyrics.
2. Parse adapted lyrics character-by-character against durations; detect "AP" markers as phrase boundaries; build timeline entries: ("AP", start, end) or ("CHAR", char, start, end).
3. Group consecutive characters between AP markers into text segments; subdivide instrumental breaks >12s into "bgm" chunks at 12s intervals.
4. Combine text and AP chunks; sort by (timestamp, content\_type\_priority --- text before AP); export sentence\_data JSON (3-decimal timestamps) to gen\_audio\_timestamps.json.

Error Handling: duration validation, timeline boundary checks, file I/O errors, encoding preservation.
\end{lstlisting}
\end{minipage}
\end{figure*}

\begin{figure*}[t]
\centering
\begin{minipage}{\textwidth}
\begin{lstlisting}[
    caption={SVCCoverist Structure},
    label=lst:svccoverist
]
Input: source_audio_path, target_vocal_path
Output: synthesized_audio_path

Algorithm:
1. Validate source/target audio paths; set output as "{source\_name}\_{target\_name}.wav" in ../../final; create output directory.
2. Navigate to tools/seed-vc; save and override PYTHONPATH for module resolution.
3. Run Seed-VC inference: \texttt{python inference.py --source <src> --target <tgt> --output <dir> --f0-condition True} (preserves F0 pitch, transfers target timbre).
4. Monitor subprocess with UTF-8/fallback decoding; validate return\_code==0 and output file creation.
5. Restore original working directory and PYTHONPATH; return output path.

Error Handling: path validation, subprocess return codes, Unicode fallback, environment restoration.
\end{lstlisting}
\end{minipage}
\end{figure*}

\begin{figure*}[t]
\centering
\begin{minipage}{\textwidth}
\begin{lstlisting}[
    caption={SVCSingle Structure},
    label=lst:svcsingle
]
Input: adapted_lyrics_string, midi_analysis_path, song_name
Output: synthesized_vocal_audio_path

Algorithm:
1. Load MIDI analysis JSON (UTF-8); extract notes, durations, and lyrics; split by "AP" markers for phrase boundaries.
2. Create segments with min 0.15s duration; aggregate short segments to 0.5s minimum via text concatenation (" | " note separator); preserve AP markers.
3. Batch synthesize via DiffSinger: generate per-segment annotation JSON ({song\_name}\_part\_{start}-{end}.json); output WAV files.
4. Apply phase vocoder time stretching (librosa, 44.1kHz): stretch factor = current/target duration; sample-level align (truncate/pad).
5. Consolidate: convert to 16-bit PCM, normalize (np.clip * 32767), insert silence for AP markers, concatenate segments.
6. Export to dataset/mad\_svc/cover/; cleanup temp files.

Error Handling: JSON/audio validation, duration mismatch correction, sample rate consistency.
\end{lstlisting}
\end{minipage}
\end{figure*}

\begin{figure*}[t]
\centering
\begin{minipage}{\textwidth}
\begin{lstlisting}[
    caption={StandUpConversion Structure},
    label=lst:standupconversion
]
Input: segment_directory, metadata_path
Output: timestamp_json_path

Algorithm:
1. Load metadata JSON (tone, text, reaction indicators); validate segment directory.
2. For each metadata entry, load {seg\_dir}/{index}.wav via soundfile; compute duration (samples/sample\_rate); track cumulative timing.
3. Generate chunk data: {id (1-based), timestamp (3-decimal), content}; preserve sequential ordering.
4. Verify timestamp progression and total duration; export sentence\_data JSON to timestamps.json (UTF-8, ensure\_ascii=False).

Error Handling: audio file validation, sample rate consistency, metadata correlation, JSON export errors.
\end{lstlisting}
\end{minipage}
\end{figure*}

\begin{figure*}[t]
\centering
\begin{minipage}{\textwidth}
\begin{lstlisting}[
    caption={CrossTalkConversion Structure},
    label=lst:crosstalkconversion
]
Input: segment_directory, metadata_path
Output: timestamp_json_path

Algorithm:
1. Load metadata JSON with role info (dou\_gen/peng\_gen), tone, text, and reaction indicators; validate segment directory.
2. For each entry, load {seg\_dir}/{index}.wav via soundfile; compute duration; track cumulative timing.
3. Format content as "[{role}] {dialogue\_text}"; generate chunk data: {id, timestamp (3-decimal), content}.
4. Verify timestamp progression and total duration; export sentence\_data JSON to timestamps.json (UTF-8).

Error Handling: audio file validation, sample rate consistency, metadata correlation, role validation.
\end{lstlisting}
\end{minipage}
\end{figure*}

\begin{figure*}[t]
\centering
\begin{minipage}{\textwidth}
\begin{lstlisting}[
    caption={RhythmContentGenerator Structure},
    label=lst:videocontentextractionagent
]
Input: user_requirements, rhythm_analysis_directory, video_segments_path
Output: video_scene_path, video_summary_path

Algorithm:
1. Load video segments JSON; build caption bank: transform raw captions to numbered format ("Video Segments {id}:").
2. Feed caption bank as video context to the LLM (system: "creative beat sync video producer"); also parse rhythm\_points.json for beat count and load rhythm visualization as pacing reference.
3. Global-aware storyboard generation: maintain multi-turn conversation state; combine rhythm reference + video captions + user requirements into a structured prompt (/////-delimited output, max 2 sentences/scene, match rhythm patterns).
4. API calls via tenacity.retry (exponential backoff: min=2s, max=60s, 5 attempts); append conversation history for continuity.
5. Export video\_scene.json (UTF-8) with user\_idea, video\_summary, and segment\_scene fields.

Error Handling: retry on API failures, fallbacks for missing rhythm/video files, truncation for oversized content.
\end{lstlisting}
\end{minipage}
\end{figure*}

\clearpage

\begin{figure*}[t]
\centering
\begin{minipage}{\textwidth}
\begin{lstlisting}[
    caption={NewsContentGenerator Structure},
    label=lst:newscontentgenerator
]
Input: user_requirements, news_presentation_style, video_directory
Output: video_scene_json

Algorithm:
1. Scan for .lab transcript files (multi-encoding: utf-8, gb18030, gbk, etc.); parse [start end text] format; load presentation style.
2. Presenter agent: adapt transcripts to narration following user ideas and style (third-person, min 11 words/sentence, remove section numbers, expand abbreviations).
3. Judger agent: reformat into /////-delimited segments per sentence; remove commas.
4. Generate English visual-scene keywords per segment (max 1 sentence, preserve proper nouns); count sections; export video\_scene.json (UTF-8).

Error Handling: encoding fallbacks, API retry (5 attempts, exponential backoff), 15K char input truncation, judger fallback.
\end{lstlisting}
\end{minipage}
\end{figure*}

\begin{figure*}[t]
\centering
\begin{minipage}{\textwidth}
\begin{lstlisting}[
    caption={Video Event Overview Presentation Style},
    label=lst:video_event_presentation_style
]
Core: third-person narrative, factual accuracy (no fabricated specs), proper subject attribution, no greetings/farewells.
Opening (first 3 sentences): compelling hook establishing central tension/innovation; approaches: situational, hypothetical, generational, or technical revelation.
Body: timeline-based chronological progression with clear thread; avoid generalizations and banned phrases ("The event shows...", "reveals...", "demonstrates...", word "event" in body); use accessible language; tight pacing and smooth transitions.
Closing: thematic elevation without summarization; refine deeper meaning; connect to broader implications.
QA: grammar, third-person consistency, factual verification, character attribution, banned-phrase compliance, timeline coherence.
\end{lstlisting}
\end{minipage}
\end{figure*}

\begin{figure*}[t]
\centering
\begin{minipage}{\textwidth}
\begin{lstlisting}[
    caption={CommentaryContentGenerator Structure},
    label=lst:commentarycontentgenerator
]
Input: user_requirements, source_text_path, commentary_presentation_style
Output: video_scene_json

Algorithm:
1. Load source text (multi-encoding: utf-8, gb18030, gbk, etc.; 30K char limit).
2. Presenter agent: adapt source to narration following user ideas and presentation style (source language, max 3 commas/sentence, preserve original dialogues, clear narrative flow).
3. Judger agent: segment into /////-delimited sentences; remove chapter numbers and extra punctuation.
4. Visual scene generation: convert to English visual-scene descriptions (max 1 sentence/section, describe character appearances, keep ///// markers).
5. Export video\_scene.json (UTF-8) with reqs, content\_created, segment\_scene.

Error Handling: encoding fallbacks, API retry (5 attempts, exponential backoff), 30K truncation.
\end{lstlisting}
\end{minipage}
\end{figure*}

\begin{figure*}[t]
\centering
\begin{minipage}{\textwidth}
\begin{lstlisting}[
    caption={Content Writing Methodology},
    label=lst:content_writing_methodology
]
Opening: first 3 sentences with compelling hook. Approaches include identity immersion, knowledge-guided, hypothetical scenarios, role-playing, time-travel hypothetical, real-event revelation, ability paradox, expectation reversal, extreme situations, and suspenseful questions.
Main Body: timeline-based plot progression, clear narrative thread, max 3 commas/sentence, smooth transitions, key character dialogue.
Prohibited: literary analysis, thematic summaries, phrases like "The story shows...", "demonstrates...", word "story" in body.
Required Transitions: high-frequency use of "but", "yet", "however", "turns out", "suddenly", "unexpectedly", "finally", etc.
Narrative Techniques: short sentences for pacing, sensory/emotional adjectives, contrast/emphasis, conversational tone, rhetorical devices (metaphor, hyperbole), strategic suspense, dramatic contrast.
Closing: extract deeper meaning, elevate themes; avoid "story ending"/"story beginning".
\end{lstlisting}
\end{minipage}
\end{figure*}

\begin{figure*}[t]
\centering
\begin{minipage}{\textwidth}
\begin{lstlisting}[
    caption={TTSWriter Structure},
    label=lst:ttswriter
]
Input: user_requirements, audio_path
Output: rewritten_speech_path

Algorithm:
1. Scan slice directory for .lab files; aggregate transcripts + original transcript (UTF-8); build structured format per slice.
2. Claude LLM: creative recreation --- preserve linguistic style and sentence structure, replace specific content, length variation <=2 chars.
3. Format template: "{index}. Original: {original}\n   Recreation:"; apply user requirements; enforce length constraints.
4. DeepSeek LLM: extract recreation content line by line; save raw\_speech.txt and final speech.txt (UTF-8).

Error Handling: lab file validation, UTF-8 encoding, LLM API failures, directory structure validation.
\end{lstlisting}
\end{minipage}
\end{figure*}

\begin{figure*}[t]
\centering
\begin{minipage}{\textwidth}
\begin{lstlisting}[
    caption={SVCAdapter Structure},
    label=lst:svcadapter
]
Input: user_requirements, midi_analysis_path, song_name
Output: adapted_lyrics_string

Algorithm:
1. Load MIDI analysis JSON (UTF-8); extract original lyrics; parse structure by "AP" markers into LYRICS placeholders and AP delimiters.
2. Generate numbered template per segment: "{index}. Original: {content}\n   Character limit: {count}\n   Recreation:".
3. Claude LLM: recreate lyrics following character limits, rhyme/rhythm, and narrative flow.
4. DeepSeek LLM: extract recreated lyrics line by line.
5. Character count alignment (max 5 retries): context-aware correction via Claude; fallback padding ("la") or truncation.
6. Reconstruct: merge aligned lyrics with AP markers; export raw\_lyrics.txt, lyrics.txt, script.txt, and {song\_name}\_cover.json.

Error Handling: JSON validation, character count mismatch correction, LLM API failures.
\end{lstlisting}
\end{minipage}
\end{figure*}

\begin{figure*}[t]
\centering
\begin{minipage}{\textwidth}
\begin{lstlisting}[
    caption={StandUpAdapter Structure},
    label=lst:standupadapter
]
Input: user_requirements, reference_script_path
Output: segmented_standup_script

Algorithm:
1. Load reference script (UTF-8); define tone markers ([Natural], [Confused], [Empathetic], [Exclamatory]) and atmosphere cues ([Laughter], [Cheers]).
2. Claude LLM: adapt into stand-up comedy --- each line begins with tone marker, atmosphere cues at key punchlines (3-4 total), preserve humor while localizing references, target 3-5 min performance.
3. Structure: "# title\n[Tone]content...\n[Tone]content...[Atmosphere]\n..."; save to stand-up.txt.

Error Handling: file validation, Claude API failures, UTF-8 encoding.
\end{lstlisting}
\end{minipage}
\end{figure*}

\begin{figure*}[t]
\centering
\begin{minipage}{\textwidth}
\begin{lstlisting}[
    caption={StandUpSynth Structure},
    label=lst:standupsynth
]
Input: segmented_script, target_vocal_directory, reaction_directory
Output: merged_audio_path, segment_directory, metadata_path

Algorithm:
1. Load CosyVoice2-0.5B (load\_jit/load\_trt=False, fp16=False); set up vocal and reaction directories.
2. Split script by lines, skip title; DeepSeek LLM parses each line -> JSON {tone, text, reaction}; validate and strip markdown from response.
3. Per segment: CosyVoice2 zero-shot synthesis using tone-specific prompts ({tone}.lab + {tone}.wav); save {index}.wav.
4. If reaction present, load {reaction}.wav; concatenate speech + reaction via PyDub AudioSegment.
5. Merge all segments (0.wav..cnt.wav) to ../final/stand\_up.wav; export metadata as stand-up.json.

Error Handling: model loading, JSON parsing, missing reaction files, audio validation, environment restoration.
\end{lstlisting}
\end{minipage}
\end{figure*}

\clearpage

\begin{figure*}[t]
\centering
\begin{minipage}{\textwidth}
\begin{lstlisting}[
    caption={CrossTalkAdapter Structure},
    label=lst:crosstalkladapter
]
Input: user_requirements, reference_script_path, dou_gen_directory, peng_gen_directory
Output: segmented_crosstalk_script

Algorithm:
1. Extract performer names (dou\_gen/peng\_gen) from directory basenames; load reference script (UTF-8).
2. Claude LLM: adapt into crosstalk dialogue --- role-alternating lines with [Natural]/[Confused]/[Emphatic] tone markers (no consecutive same tone >2 lines), localize humor for Chinese audience, balanced dialogue allocation.
3. Format: "Title\n[tone] Role: content...\n..."; save to cross-talk.txt.

Error Handling: file validation, Claude API failures, UTF-8 encoding.
\end{lstlisting}
\end{minipage}
\end{figure*}

\begin{figure*}[t]
\centering
\begin{minipage}{\textwidth}
\begin{lstlisting}[
    caption={CrossTalkSynth Structure},
    label=lst:crosstalksynth
]
Input: segmented_script, funny_guy_directory, setup_guy_directory
Output: merged_audio_path, segment_directory, metadata_path

Algorithm:
1. Extract role names (dou\_gen/peng\_gen) from directories; load CosyVoice2-0.5B.
2. Split script by lines; DeepSeek LLM parses each line -> JSON {role, tone, text, reaction}; validate performer names.
3. Per segment: CosyVoice2 zero-shot synthesis using role/tone-specific prompts ({role}/{tone}.lab + .wav at 16kHz); save {index}.wav.
4. Merge all segments via PyDub to ../final/cross\_talk.wav; export metadata as cross-talk.json.

Error Handling: model loading, JSON parsing, missing prompt files, role validation, environment restoration.
\end{lstlisting}
\end{minipage}
\end{figure*}

\begin{figure*}[t]
\centering
\begin{minipage}{\textwidth}
\begin{lstlisting}[
    caption={VideoConversion Structure},
    label=lst:videoconversion
]
Input: timestamp_path
Output: video_scene_path

Algorithm:
1. Load timestamp JSON (cut\_points.json); extract content from sentence\_data.chunks; join with ///// separators.
2. LLM (DeepSeek): generate English visual-scene descriptions --- keep ///// markers, max 1 sentence/segment, describe character appearances, don't directly translate.
3. Export video\_scene.json (UTF-8) with segment\_scene and content\_created; validate segment count matches input.

Error Handling: file/JSON validation, empty segments, LLM parsing fallbacks.
\end{lstlisting}
\end{minipage}
\end{figure*}

\begin{figure*}[t]
\centering
\begin{minipage}{\textwidth}
\begin{lstlisting}[
    caption={VideoQA Structure with System Prompt},
    label=lst:videoqa
]
Input: video_directory, output_path, save_history_flag
Output: transcript_path, qa_session_results

Algorithm:
1. Load Whisper model; scan directory for videos; transcribe each (30s chunks, batch-optimized) with markers ("=== VIDEO: filename ===").
2. Consolidate transcripts; truncate at 50K chars; initialize interactive Q&A session.
3. Q&A loop: answer user questions strictly from transcripts with source video citation; state if answer not found; log conversations with timestamps.
4. Save Q&A history and return results.

Error Handling: transcription fallback, graceful session termination, comprehensive logging.
\end{lstlisting}
\end{minipage}
\end{figure*}

\begin{figure*}[t]
\centering
\begin{minipage}{\textwidth}
\begin{lstlisting}[
    caption={VideoSummarizationGenerator Structure},
    label=lst:videosummarizationgenerator
]
Input: user_idea, video_directory, presentation_style_path, output_path
Output: content_output, status, processed_videos, transcript_source

Algorithm:
1. Load whisper-large-v3-turbo ASR pipeline (max\_new\_tokens=128, chunk\_length\_s=30, batch\_size=16); auto-detect CUDA/CPU.
2. Scan for video files; transcribe each with retry (3 attempts); combine with "=== VIDEO: filename ===" markers; also support text file input (multi-encoding).
3. Presenter agent: generate summarization following user ideas and presentation style (third-person, fewer bullet points, word count adherence); truncate input at 15K chars.
4. Save output as plain text; return metadata (processed\_videos, transcript\_source, status).

Error Handling: retry logic (3-5 attempts, exponential backoff), CUDA fallback, encoding fallbacks.
\end{lstlisting}
\end{minipage}
\end{figure*}

\begin{figure*}[t]
\centering
\begin{minipage}{\textwidth}
\begin{lstlisting}[
    caption={System Prompt of Intent Analysis},
    label=lst:intentanalysis1
]
You are an intent analyst.  
I will provide a set of candidate intents and a user's requirements.
Please select as many relevant candidate intents as possible that match the user's requirements. Consider factors such as keywords, audio types (speech, song, music, etc.), and other relevant dimensions.  

Candidate intents:  
{intents}  

User requirements:  
{reqs}  

Please Only output pure List format:
['intent1', 'intent2', ...]

Note! Don't output any analysis and explanations!
\end{lstlisting}
\end{minipage}
\end{figure*}

\begin{figure*}[t]
\centering
\begin{minipage}{\textwidth}
\begin{lstlisting}[
    caption={System Prompt of Intent Analysis},
    label=lst:intentanalysis2
]
You are an intent analyst.  
Previous analysis attempt failed with the following reflection:
{reflection}

Previous selected intents:
{previous_intents}

Please re-analyze the user's requirements with the candidate intents below, considering the reflection.
Select as many relevant candidate intents as possible.

Candidate intents:  
{intents}  

User requirements:  
{reqs}  

Please Only output pure List format:
['intent1', 'intent2', ...]

Note! Don't output any analysis and explanations!
\end{lstlisting}
\end{minipage}
\end{figure*}

\begin{figure*}[t]
\centering
\begin{minipage}{\textwidth}
\begin{lstlisting}[
    caption={System Prompt of Agent Graph},
    label=lst:agentgraphprompt
]
You are an Agent Graph Designer. I will provide User Requirement and Registered Agents (Name, Description and Parameters information):

Your task is to:
1. Judge Feasibility:
   - Evaluate implementation feasibility of User Requirements
   - Output: "Feasible" or "Infeasible" (strictly one of these)
2. Design Executable Agent Graph:
   - Format: List
   - Agent Graph shall contain metadata for each Agent Node including:
     * name: (string)
     * inputs: (list of input parameter objects with):
     * parameter: input parameter name
     * description: brief parameter description
     * outputs: (list of output parameter objects with):
     * parameter: output parameter name
     * description: brief parameter description
     * links: (list of dictionaries) where each dictionary specifies:
         - key of dictionaries: target agent name
         - value of dictionaries: target agent's input parameter name that this output connects to
3. Generate Agent Chain:
   - Format: List
   - Generate the Agent Chain based on the description of the Agent and the sequential information contained in the designed Agent Graph
4. Generate User Input Graph
   - Format: List
   - Parameter nodes with no in-degree (no incoming edges) are uniformly considered to require user input.
   - Parameter nodes with no in-degree may have different names but share the same user input, meaning a single user input parameter can point to multiple such nodes.
   - Parameter nodes with no in-degree that are linked to user input should be represented in the format **AgentName.input_parameter**
   - Generate the User Input Graph based on the Agent descriptions and parameter passing information in the designed Agent Graph.
5. Output Reasoning:
   - If Feasible, Provide concise reasoning (<200 words) explaining the entire workflow logic
   - If Infeasible, Specify exact failure reasons (<200 words)

In addition to the above formatting requirements, please also note the following:
1. For each element of **outputs** in each Agent Node:
   - Ensure that the **links** in the **outputs** point to an input parameter that actually exists in the next Agent Node.
   - The output parameter name does not need to match the input parameter name in the next Agent Node.
   - Ensure the output parameter's description and type match the input parameter requirements of the next Agent Node. For example, a file path output cannot be passed to a directory path input.
2. Final JSON Output Format Specification:
{
    "Feasibility": "Feasible" or "Infeasible",
    "Agent Graph": ...,
    "Agent Chain": ...,
    "User Input Graph": ...,
    "Reasoning": ...
}
Strictly follow JSON output format!
\end{lstlisting}
\end{minipage}
\end{figure*}

\clearpage

\begin{figure*}[t]
\centering
\begin{minipage}{\textwidth}
\begin{lstlisting}[
    caption={System Prompt for Initial Evaluation},
    label=lst:initialevaluation
]
You are an agent graph validation system.
I will provide:
1. User Requirement
2. Registered agent metadata
3. Candidate Agent Graph
4. An Agent Chain derived from the Candidate Agent Graph
5. Required User Inputs

User Requirements:
{reqs}

Metadata of registered agents:
{tools}

Task: Evaluate the candidate agent graph:

Candidate Agent Graph:
{agent_graph}

Agent Chain:
{agent_chain}

Required User Inputs:
{user_inputs}

Evaluation Criteria:
1. Based on the Metadata of registered agents and parameter passing in the Agent Graph, determine from multiple aspects whether the user requirements can be fulfilled:
   - Execution sequence of agents in the Agent Graph
   - For parameter nodes with no incoming edges, they are uniformly considered as user inputs, but it is necessary to determine whether they should be provided by the user or by the parent agent
   - Validate that the necessary output parameters are correctly routed to the intended agent and the expected input parameters.
2. There should be no functionally redundant agents (e.g., repeatedly adding audio tracks to a video).
3. For vaguely mentioned requirements in user needs, lenient evaluation is acceptable. For example, if the user requests audio quality improvement, it's sufficient as long as at least one relevant agent in the graph meets this requirement.

Please Only output pure JSON format:
{{
"Result": '0' if correct else '1',
"Reasoning": Concisely state the key reasons why a score of '0' or '1' was assigned (<100 words).
}}
\end{lstlisting}
\end{minipage}
\end{figure*}

\begin{figure*}[t]
\centering
\begin{minipage}{\textwidth}
\begin{lstlisting}[
    caption={System Prompt for Secondary Evaluation},
    label=lst:secondaryevaluation
]
You are an agent graph reflection system.
I will provide:
1. User Requirement
2. Registered agent metadata
3. Candidate Agent Graph
4. An Agent Chain and User Input Graph derived from the Candidate Agent Graph
5. Previous validation result 

User Requirements:
{reqs}

Metadata of registered agents:
{tools}

Task: Evaluate the candidate agent graph:

Candidate Agent Graph:
{agent_graph}

Agent Chain:
{agent_chain}

Required User Input Graph:
{user_inputs}

Previous validation result:
{judge_res}

Reflection Task:
1. If the previous validation result is '0', please reflect on whether there were any overlooked aspects based on the **Evaluation Criteria** and the reasoning behind the previous validation result.
2. If the previous validation result is '1', please reflect on whether the reasoning behind the previous validation result was correct.

Evaluation Criteria:
1. Based on the Metadata of registered agents and parameter passing in the Agent Graph, determine from multiple aspects whether the user requirements can be fulfilled:
   - Execution sequence of agents in the Agent Graph
   - For parameter nodes with no incoming edges, they are uniformly considered as user inputs, but it is necessary to determine whether they should be provided by the user or by the previous agent
   - Validate that the necessary output parameters are correctly routed to the intended agent and the expected input parameters.
   - Validate that the output parameters' description and type match the input requirements of the next agent.
   - Not all output parameters are necessarily mapped to the input requirements of the next agent. Redundant output parameters may exist, but they should not interfere with the fulfillment of user requirements.
2. There should be no functionally redundant agents (e.g., repeatedly adding audio tracks to a video).
3. For vaguely mentioned requirements in user needs, lenient evaluation is acceptable. For example, if the user requests audio quality improvement, it's sufficient as long as at least one relevant agent in the graph meets this requirement.
 
Please Only output pure JSON format:
{{
"Result": '0' if correct else '1',
"Reasoning": Concisely state the key reasons why a score of '0' or '1' was assigned (<100 words).
}}
\end{lstlisting}
\end{minipage}
\end{figure*}

\begin{figure*}[t]
\centering
\begin{minipage}{\textwidth}
\begin{lstlisting}[
    caption={User Requirement of Audio Preprocessing},
    label=lst:reqaudiopreprocessing
]
User Requirement: I need to extract the audio from a video file, improve its sound quality by normalizing the loudness, and then mix it with some background music. The result should be a clean audio track that balances the original audio with background music.
\end{lstlisting}
\end{minipage}
\end{figure*}

\begin{figure*}[t]
\centering
\begin{minipage}{\textwidth}
\begin{lstlisting}[
    caption={User Requirement of Audio Preprocessing},
    label=lst:audiopreprocessing
]
{
"Feasibility": "Feasible",
"Intent List": ["Loudness Normalization", "Add BGM", "Audio Optimization"],
"Agent Graph": [
    {"node": "LoudnessNormalizer", "inputs": [{"name": "data_dir", "description": "Directory of audio files to be normalized"}], "outputs": [{"name": "status", "description": "Execution status (success/error)", "links": []}]},
    {"node": "Mixer", "inputs": [{"name": "bgm_path", "description": "Path to the BGM file"}, {"name": "audio_path", "description": "Audio to be mixed into the background music"}], "outputs": [{"name": "audio_path", "description": "Path to the synthesized audio", "links": []}]}
],
"Agent Chain": ["LoudnessNormalizer", "Mixer"],
"User Input Graph": [
    {"node": "raw_audio_directory", "description": "Directory containing the user's raw audio recording that needs to be normalized", "links": [{"LoudnessNormalizer": "data_dir"}]},
    {"node": "normalized_audio_path", "description": "Path to the normalized audio file after processing by LoudnessNormalizer", "links": [{"Mixer": "audio_path"}]},
    {"node": "background_music_path", "description": "File path to the background music to be mixed with the voice recording", "links": [{"Mixer": "bgm_path"}]}
],
"Reasoning": "This workflow normalizes the user's raw audio recording and then mixes it with background music. First, LoudnessNormalizer processes the raw audio files to ensure consistent volume levels. Since LoudnessNormalizer doesn't output file paths (only a status), we require the user to provide the path to the normalized audio file after processing. Then, the Mixer agent combines this normalized audio with background music, ensuring the music doesn't overpower the voice. The workflow addresses the requirement to add background music to a recording while maintaining proper audio balance."
}
\end{lstlisting}
\end{minipage}
\end{figure*}

\begin{figure*}[t]
\centering
\begin{minipage}{\textwidth}
\begin{lstlisting}[
    caption={User Requirement of Storytelling Video},
    label=lst:reqstorytelling
]
User Requirement: I want to create a commentary video about a novel. The video should have professional voiceover with background music and appropriate visuals that match the commentary content.
\end{lstlisting}
\end{minipage}
\end{figure*}

\begin{figure*}[t]
\centering
\begin{minipage}{\textwidth}
\begin{lstlisting}[
    caption={Case of Storytelling Video},
    label=lst:casestorytelling
]
{
"Feasibility": "Feasible",
"Intent List": ["Text-to-Speech", "Commentary", "Add BGM", "Video Edit"],
"Agent Graph": [
    {"node": "CommentaryContentGenerator", "inputs": [{"name": "reqs", "description": "User's idea for the commentary video including word count requirements"}, {"name": "source_text", "description": "File path to the novel source text"}, {"name": "comm_present_style", "description": "File path to commentary presentation style for content generation"}], "outputs": [{"name": "video_scene_path", "description": "File path storing scene semantics for video storyboard sound synthesis.", "links": [{"VoiceGenerator": "video_scene_path"}, {"VideoSearcher": "video_scene_path"}]}]},
    {"node": "VoiceGenerator", "inputs": [{"name": "video_scene_path", "description": "Path to a custom scene JSON file"}, {"name": "target_vocal_path", "description": "Path to the target timbre for voice generation"}], "outputs": [{"name": "audio_path", "description": "Path to the synthesized audio", "links": [{"Mixer": "audio_path"}]}, {"name": "timestamp_path", "description": "Path to video frame timestamp", "links": [{"VideoEditor": "timestamp_path"}]}]},
    {"node": "Mixer", "inputs": [{"name": "bgm_path", "description": "Path to the BGM file"}, {"name": "audio_path", "description": "Audio to be mixed into the background music"}], "outputs": [{"name": "audio_path", "description": "Path to the synthesized audio", "links": [{"VideoEditor": "audio_path"}]}]},
    {"node": "VideoPreloader", "inputs": [{"name": "video_dir", "description": "Directory containing the source MP4 video files to be processed"}], "outputs": [{"name": "status", "description": "Execution status (success/error)", "links": []}]},
    {"node": "VideoSearcher", "inputs": [{"name": "video_scene_path", "description": "File path storing scene semantics for video storyboard sound synthesis."}], "outputs": [{"name": "status", "description": "Execution status (success/error)", "links": []}]},
    {"node": "VideoEditor", "inputs": [{"name": "video_dir", "description": "Directory containing source video files"}, {"name": "audio_path", "description": "Path to the audio"}, {"name": "timestamp_path", "description": "JSON File path used to store and load the timestamp of the end of each video segment"}], "outputs": [{"name": "video_path", "description": "Path to the generated video file", "links": []}]}
],
"Agent Chain": ["CommentaryContentGenerator", "VoiceGenerator", "Mixer", "VideoPreloader", "VideoSearcher", "VideoEditor"],
"User Input Graph": [
    {"node": "commentary_requirements", "description": "User's specific requirements for the commentary video including word count and style", "links": [{"CommentaryContentGenerator": "reqs"}]},
    {"node": "novel_file", "description": "File path to the novel text that will be the subject of commentary", "links": [{"CommentaryContentGenerator": "source_text"}]},
    {"node": "commentary_style", "description": "File path to the desired presentation style for the commentary", "links": [{"CommentaryContentGenerator": "comm_present_style"}]},
    {"node": "voice_timbre", "description": "Path to the target vocal timbre file for professional voiceover", "links": [{"VoiceGenerator": "target_vocal_path"}]},
    {"node": "background_music", "description": "Path to the background music file", "links": [{"Mixer": "bgm_path"}]},
    {"node": "visuals_directory", "description": "Directory containing source video files for visual content", "links": [{"VideoPreloader": "video_dir"}, {"VideoEditor": "video_dir"}]}
],
"Reasoning": "The workflow begins with generating commentary content about the novel based on user requirements. This content is structured with scene semantics. The VoiceGenerator then creates professional voiceover audio and provides timestamps for visual syncing. In parallel, VideoPreloader prepares video files from the user's library while VideoSearcher identifies appropriate visual clips based on the commentary content. The voiceover is mixed with background music by the Mixer agent. Finally, VideoEditor combines the mixed audio track with matching visuals using the timestamps to synchronize content, producing a complete commentary video about the novel with professional narration, appropriate background music, and relevant visuals."
}
\end{lstlisting}
\end{minipage}
\end{figure*}

\begin{figure*}[t]
\centering
\begin{minipage}{\textwidth}
\begin{lstlisting}[
    caption={MLLM Video Clip Matching Prompt with Frame Integration},
    label=lst:video_clip_matching_prompt
]
Prompt Construction with Frame Integration:

1. Text prompt construction:
prompt = f"""
Here is a video broken down into {len(chunks)} clips of 3 seconds each. 
Each clip shows multiple frames from that time segment.
The clips are structured as follows:
"""

for chunk in chunks:
    prompt += f"Clip {chunk['chunk_idx']}\n"

prompt += f"""
Identify {num_periods} specific clips that best correspond to the following captions:
"""

for i, caption in enumerate(shuffled_captions):
    prompt += f"{i+1}. {caption}\n"

prompt += f"""
Response with a JSON structure containing the clip numbers that best match each caption. 

Rules:
1. Each clip_id should be a number (not a string)
2. Include exactly {num_periods} selections
3. Choose from the clip numbers shown (0 to {len(chunks)-1})
4. Keep the reason concise (1-2 sentences)
5. Ensure your output is valid JSON - no trailing commas, proper quotes, etc.
6. Do not include the ```json prefix or ``` suffix around your response
7. Do not answer anything unrelated

Return only the JSON object with no additional text.

Output Example:
{{
"selections": [
    {{
    "caption": "[caption text]",
    "clip_id": [clip number],
    "reason": "[brief description of what is seen in this clip]"
    }},
    ...
]
}}
"""

2. Frame integration into multimodal message structure:
messages = [{"role": "user", "content": []}]

# Add text prompt as first content element
messages[0]["content"].append({"type": "text", "text": prompt})

# Sequential frame insertion for visual analysis
for img_base64 in frames:
    messages[0]["content"].append({
        "type": "image_url", 
        "image_url": {
            "url": f"data:image/jpeg;base64,{img_base64}"
        }
    })
\end{lstlisting}
\end{minipage}
\end{figure*}

\end{document}